\documentclass{article}
\usepackage{iclr2027_conference,times}
\usepackage{microtype}

\usepackage{amsmath,amsfonts,bm}

\def\eqref#1{equation~\ref{#1}}

\def\1{\bm{1}}

\DeclareMathAlphabet{\mathsfit}{\encodingdefault}{\sfdefault}{m}{sl}
\SetMathAlphabet{\mathsfit}{bold}{\encodingdefault}{\sfdefault}{bx}{n}

\usepackage{amssymb}
\usepackage{algorithm}
\usepackage{algpseudocode}
\usepackage{graphicx}
\usepackage{booktabs}
\usepackage{colortbl}
\usepackage{multirow}
\usepackage{float}
\usepackage{wrapfig}
\usepackage{needspace}
\usepackage[most]{tcolorbox}
\usepackage{url}
\usepackage{hyperref}
\hypersetup{
    colorlinks=true,
    citecolor=blue,
    linkcolor=blue,
    urlcolor=blue
}

\title{UniSkill: Learning Actor-Aligned Skill \\
Proposals for an Evolving Policy}

\author{
\normalfont
{\small
\textbf{Yifei Lu}$^{1,2}$\thanks{Equal Contributions. E-mail: okiilu.lyf@qiantangcredit.com, chengliu2@link.cuhk.edu.cn},
\textbf{Cheng Liu}$^{3}$\footnotemark[1],
\textbf{Dianzhi Yu}$^{3}$,
\textbf{Hui Xiang}$^{1,2}$,
\textbf{Ji Zhang}$^{1,2}$,
\textbf{Yuanchu Xiao}$^{1,2}$,
\textbf{Rong Liang}$^{1,2}$\thanks{Corresponding author.}}\\[0.35em]
$^{1}$Qiantang Credit, Hangzhou, China
\hspace{1.5em}$^{2}$Ant Group, Hangzhou, China\\[0.2em]
$^{3}$The Chinese University of Hong Kong
}

\iclrfinalcopy

\begin{document}

\raggedbottom
\setlength{\textfloatsep}{12pt plus 2pt minus 2pt}
\setlength{\intextsep}{8pt plus 2pt minus 2pt}

\maketitle
\ificlrfinal
\fancyhf{}
\renewcommand{\headrulewidth}{0pt}
\fi

\begin{abstract}
Large language model agents can improve across tasks by retaining reusable
skills distilled from prior interactions. Recent work jointly optimizes task
execution and skill extraction, enabling the policy and skillbank to co-evolve.
However, as the actor continues learning, rewarding skill proposals through
their reuse in subsequent training steps may conflate skill benefits with actor
improvement, while directly testing each proposed skill requires costly additional
actor rollouts. In this paper, we introduce UniSkill, which uses a shared policy
to interact with the environment and propose skillbank edits
(\textsc{Add}, \textsc{Update}, or \textsc{No Edit}) from the resulting
trajectories. Specifically, the actor learns from environment rewards, while
contrastive action feedback guides skill proposal learning. This feedback
provides an actor-alignment signal by measuring how replacing the retrieved
skill with a proposed skill changes the current actor's action log-likelihood
gap between previously collected successful and failed trajectories from the
same task, thereby avoiding new rollouts for each proposal. Since proposal-level
feedback may suppress an otherwise appropriate edit operation when the proposed
skill content scores poorly, we further apply skill-edit support regularization
to preserve exploration. Empirically, UniSkill achieves strong
performance, reaching 98.4\% success on ALFWorld and 84.7\% on WebShop while
maintaining stable joint training. Further ALFWorld experiments show that
UniSkill remains effective when the shared policy uses a smaller backbone.
Our implementation is available at \url{https://github.com/LimOkii/UniSKill}.
\looseness=-1
\end{abstract}

\begingroup
\setlength{\intextsep}{0pt}
\begin{figure}[H]
    \centering
    \setlength{\abovecaptionskip}{2pt}
    \setlength{\belowcaptionskip}{0pt}
    \includegraphics[width=0.86\linewidth]{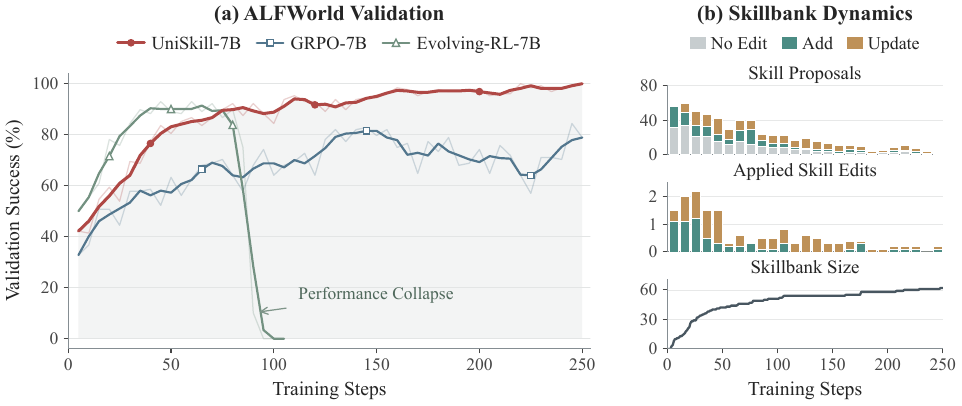}
    \caption{ALFWorld training dynamics.
    (a) UniSkill continues improving later in training, whereas Evolving-RL
    collapses after strong early gains.
    (b) The learned skillbank grows rapidly early and continues evolving through
    occasional additions and updates (non-overlapping 10-step means).
    \looseness=-1}
    \label{fig:alfworld-learning-dynamics}
\end{figure}
\endgroup
\section{Introduction}

LLM agents increasingly reuse information acquired from earlier interactions
rather than treating every task as an isolated episode
\citep{xu2025amem,ouyang2026reasoningbank,zhang2026ace,ni2026trace2skill}.
Prior work retains experience through verbal reflections, induced insights,
and executable programs
\citep{shinn2023reflexion,zhao2024expel,wang2024voyager}.
Recent work distills interaction trajectories into reusable textual skills
and stores them in a skillbank, from which relevant skills are retrieved to guide
subsequent task execution \citep{wu2026evolver,xia2026skillrl}.

Existing approaches extract skills from interaction trajectories by prompting
a powerful LLM with frozen parameters \citep{ni2026trace2skill,xia2026skillrl}.
Beyond fixed-model skill extraction, several methods jointly train the skill
proposer and the actor
\citep{muhtar2026complementary,shi2026skill1,fan2026evolvingrl}.
The actor interacts with the environment to complete tasks, and the resulting
trajectories are used to extract skills that in turn guide subsequent
interactions.  As skill extraction
becomes a learned capability, a central question is what learning feedback a
newly proposed skill should receive.
To provide this feedback, existing methods train models to generate reflections
or extract skills using task-outcome prediction accuracy or rewards from the
trajectories used for skill extraction
\citep{zhang2026retroagent,shi2026skill1}.
These signals do not assess how the current actor would behave when given
the proposed skill.  More direct feedback can be obtained by evaluating task
performance when the actor uses the skills in subsequent
interactions \citep{muhtar2026complementary,he2026reskill}.

However, the usefulness of a skill depends strongly on the actor that uses it
\citep{yu2026masa}.  During joint training, the actor may already have been
updated by the time the skill is reused.  The updated actor might succeed even
without the skill, so task success alone does not establish whether the skill
would have helped the actor at the time it was proposed.  Evaluating candidate
skills before updating the actor avoids these intervening policy changes, but
requires additional environment rollouts conditioned on each candidate skill, as in
Evolving-RL \citep{fan2026evolvingrl}.  We therefore ask whether trajectories
already collected by the current actor can provide learning feedback for new
skill proposals.
\looseness=-1

\begin{figure*}[t]
    \centering
    \includegraphics[width=0.92\textwidth]{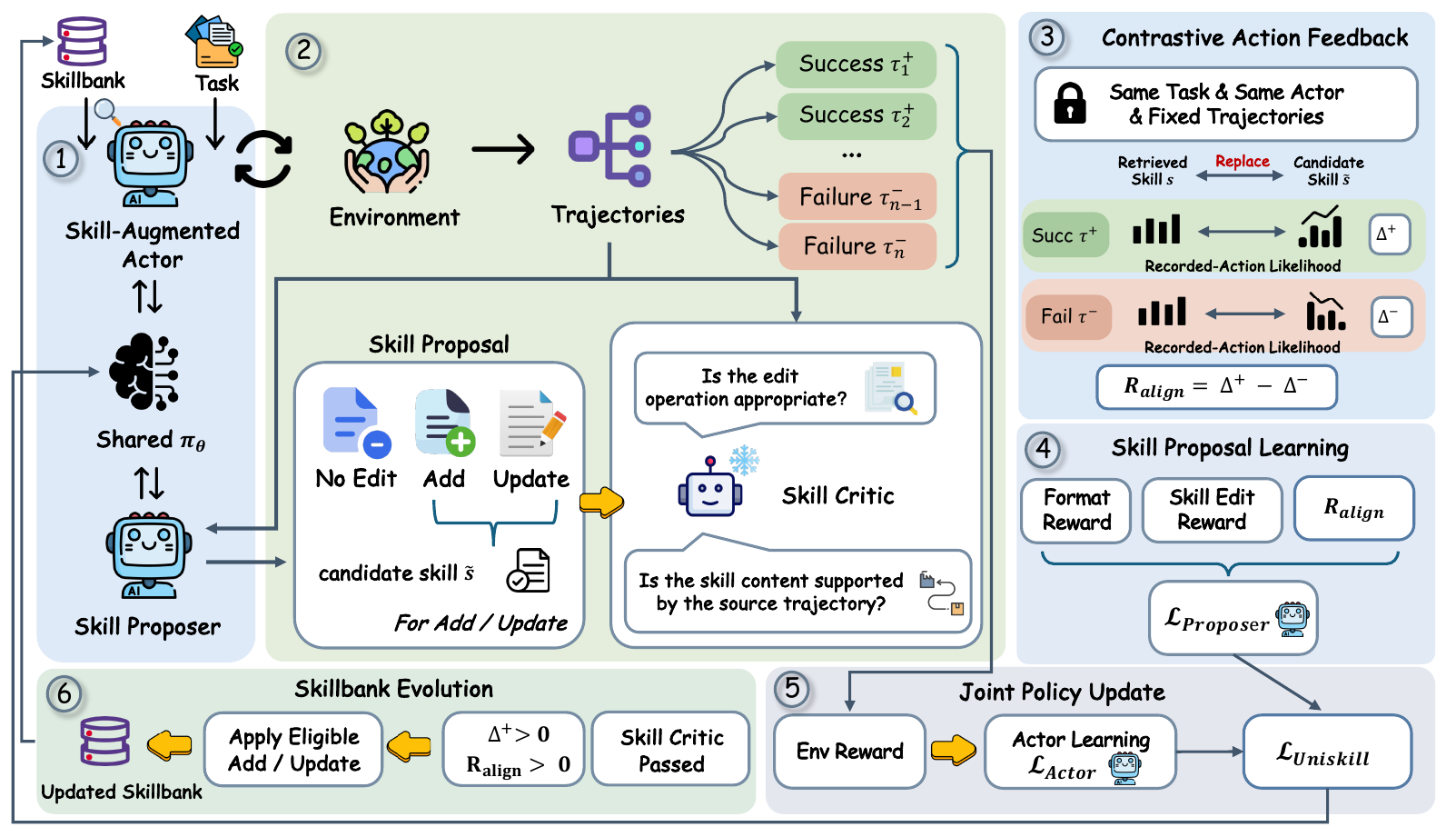}
    \caption{Overview of UniSkill.  (1--2) The shared policy interacts with the
    environment and proposes edits from the resulting trajectories, while a
    skill critic checks edit
    appropriateness and content support.  (3) Contrastive action feedback
    measures actor alignment on fixed trajectories.  (4--5) Actor and proposer
    objectives jointly update the policy.  (6) Eligible edits update the
    skillbank.\looseness=-1}
    \label{fig:uniskill-overview}
\end{figure*}

In this paper, we introduce \textbf{UniSkill}, a shared-policy framework that
learns to act with retrieved skills and generate skill proposals from completed
trajectories, as illustrated in Figure~\ref{fig:uniskill-overview}.  Each skill proposal specifies
a skill-edit operation for adding a new skill, updating the retrieved skill,
or leaving the skillbank unchanged.  For \textsc{Add} and \textsc{Update}, the
proposal also includes the corresponding new or revised skill content.
To evaluate these proposed skills, we focus on \emph{actor alignment},
the extent to which a proposed skill improves task
performance for the current actor relative to the retrieved skill.
We construct a proxy for this utility using previously collected successful
and failed trajectories from the same task.  Specifically, before the policy
update, we hold the current actor and recorded behavior fixed and measure how
replacing the retrieved skill with the proposed skill changes the action
log-likelihood gap between these trajectories.  We then combine this
\emph{contrastive action feedback} with rewards for the appropriateness of
skill-edit operations to train the skill proposer, while the actor learns from
environment rewards.  Because the skill-edit operation and proposed content
share a proposal-level training signal, negative feedback on the content may
suppress an otherwise appropriate operation.  We therefore use
skill-edit support regularization to preserve exploration of these operations
during joint training.
Skill edits that pass both skill-critic and actor-alignment checks update the
skillbank for subsequent interactions, allowing the policy and skillbank to
co-evolve.
As shown in Figure~\ref{fig:alfworld-learning-dynamics}, UniSkill achieves further
gains in validation success during later training on ALFWorld, while continuing
to add new skills and refine existing ones.

\Needspace{5\baselineskip}
Our contributions are as follows:
\begingroup
\setlength{\leftmargini}{1.5em}
\begin{itemize}
    \item We introduce UniSkill, a shared-policy framework for joint actor
    and skill proposer learning, with skill-edit support regularization that
    helps sustain exploration of edit operations.
    \item We develop contrastive action feedback by comparing proposed and
    retrieved skills using the current actor's
    action log-likelihoods on previously collected successful and failed
    trajectories, avoiding additional rollouts for each proposal and reducing
    skill-evaluation overhead.
    \item Experiments on ALFWorld and WebShop demonstrate strong performance
    among the compared methods.  Further experiments show that
    UniSkill remains effective when both the actor and proposer use a
    smaller backbone, supporting policy--skill co-evolution across model sizes.
\end{itemize}
\endgroup

\section{Related Work}

\subsection{Experience Reuse and Skill-Augmented Agents}

Prior work retains experience as verbal reflections and induced insights
\citep{shinn2023reflexion,park2023generativeagents,zhao2024expel}, executable
programs \citep{wang2024voyager,zheng2025skillweaver}, and abstracted workflows
\citep{wang2025awm}.  A common approach prompts LLMs to extract and refine
reusable guidance in external memory, without updating model parameters.
For example, ReasoningBank prompts an LLM to extract reusable strategies from
trajectories labeled as successful or failed by an LLM judge, storing them for
subsequent retrieval \citep{ouyang2026reasoningbank}.  Dynamic Cheatsheet uses
a prompted curator to revise strategy memory \citep{suzgun2026dynamiccheatsheet},
while ACE integrates extracted lessons into structured playbooks through
incremental updates \citep{zhang2026ace}.
Beyond inference-time memory adaptation, EvolveR and SkillRL integrate
experience reuse with policy optimization, refreshing stored principles or
skills as training proceeds \citep{wu2026evolver,xia2026skillrl}.  SkillGraph
models inter-skill dependencies to support structured retrieval and
graph evolution during RL \citep{li2026skillgraph}.  As policies and their
stored experience co-evolve, a central question is how to evaluate and credit
that experience for the current actor.
\looseness=-1

\subsection{Credit Assignment for Co-Evolving Policies and Skills}

RetroAgent rewards correct task-outcome predictions during reflection
\citep{zhang2026retroagent}, while Skill1 rewards skill distillation using task
returns relative to the highest historical utility among retrieved skills
\citep{shi2026skill1}.  These rewards guide skill generation, but do not directly
evaluate whether the generated skill helps the actor.  Learning feedback for skill
generation and use can instead depend on whether the actor completes a task
successfully using skills
\citep{wang2025sage,li2026arise}.  Complementary RL trains its skill proposer
from outcomes of later skill reuse \citep{muhtar2026complementary}, while
ReSkill compares skillbank revisions across training steps and discounts older
outcomes \citep{he2026reskill}.  MASA demonstrates that skill effectiveness
varies across model backbones, highlighting the need for actor-specific skill
evaluation \citep{yu2026masa}.  During joint training, by the time a proposed
skill is used and evaluated on a subsequent task, the actor may already have
been updated.  Evolving-RL evaluates
candidate skills before the joint policy update through additional
skill-conditioned rollouts \citep{fan2026evolvingrl}.  UniSkill instead holds
the current actor and recorded trajectories fixed, comparing action
likelihoods under the retrieved and proposed skills to train the skill
proposer without additional evaluation rollouts.

\section{Methodology}
\label{sec:methodology}

\subsection{Preliminaries}

UniSkill maintains a skillbank $\mathcal{S}$ and uses a shared policy
$\pi_\theta$ in two roles: an \emph{actor} that generates environment
actions conditioned on a retrieved skill, and a \emph{skill proposer}
that generates skill-edit operations and corresponding skill content from
completed trajectories.

\paragraph{Skill-Augmented Actor.}
For an episodic interactive task $q\sim\mathcal{D}$, a retriever selects a skill
$s\in\mathcal{S}$.  At each turn $t$, the actor samples an action given
the task, retrieved skill, and interaction history:
\begingroup
\setlength{\abovedisplayskip}{11pt plus2pt minus5pt}
\setlength{\abovedisplayshortskip}{6pt plus3pt}
\setlength{\belowdisplayskip}{8pt plus2pt minus5pt}
\setlength{\belowdisplayshortskip}{6pt plus3pt minus2pt}
\begin{equation}
    a_t\sim\pi_{\theta}(\cdot\mid q,s,H_t),
    \label{eq:skill-augmented-actor}
\end{equation}
\endgroup
where $H_t=(o_1,a_1,\ldots,a_{t-1},o_t)$ denotes the interaction history up to
the current observation $o_t$.
The resulting trajectory is $\tau=(o_1,a_1,\ldots,o_T,a_T,y)$, where $T$ is
the episode length and $y$ is the terminal task outcome.

\paragraph{Trajectory-Conditioned Skill Proposal.}
Given a completed source trajectory $\tau$ and an opposite-outcome trajectory
$\tau_{\mathrm{opposite}}$ from the same task $q$ and retrieved skill $s$,
the shared policy compares the two trajectories and generates a structured
skill-edit proposal:
\begingroup
\setlength{\abovedisplayskip}{11pt plus2pt minus5pt}
\setlength{\abovedisplayshortskip}{6pt plus3pt}
\setlength{\belowdisplayskip}{8pt plus2pt minus5pt}
\setlength{\belowdisplayshortskip}{6pt plus3pt minus2pt}
\begin{equation}
    (\mathrm{op},\tilde{s})\sim
    \pi_{\theta}(\cdot\mid q,s,\tau,\tau_{\mathrm{opposite}}),
    \qquad
    \mathrm{op}\in\{\textsc{No Edit},\textsc{Add},\textsc{Update}\}.
    \label{eq:skill-proposal}
\end{equation}
\endgroup
Here, $\mathrm{op}$ specifies the skill-edit operation, while $\tilde{s}$ is the
candidate skill for \textsc{Add} or \textsc{Update}.
\textsc{No~Edit} leaves the skillbank unchanged.
\textsc{Add} proposes a new skill when the trajectory reveals reusable guidance
that warrants a separate skillbank entry.
\textsc{Update} proposes a revision to the retrieved skill when the trajectory
indicates that its guidance requires correction, refinement, or extension.

\subsection{Skill-Augmented Actor Learning}
\label{sec:actor-learning}

For each sampled task $q$, the rollout policy $\pi_{\mathrm{old}}$ generates a
group $\mathcal{T}_q$ of $G$ trajectories conditioned on the same retrieved
skill $s$, which remains fixed throughout each episode.  We train the actor
with Group Relative Policy Optimization
(GRPO)~\citep{shao2024deepseekmath} using environment rewards.
Let $R_j$ be the environment reward of trajectory $j$ in the group for task $q$.
Its group-relative advantage is
\vspace{-1pt}
\begin{equation}
    \widehat A_j
    =\frac{R_j-\mu_q}{\sigma_q+\epsilon},
    \qquad
    \mu_q=\frac{1}{G}\sum_{j=1}^{G}R_j,
    \label{eq:grpo-main}
\end{equation}
\vspace{1pt}
where $\sigma_q$ is the standard deviation of rewards within the group.
We minimize the actor loss:
\begingroup
\setlength{\abovedisplayskip}{11pt plus2pt minus5pt}
\setlength{\abovedisplayshortskip}{6pt plus3pt}
\setlength{\belowdisplayskip}{8pt plus2pt minus5pt}
\setlength{\belowdisplayshortskip}{6pt plus3pt minus2pt}
\begin{equation}
    \mathcal{L}_{\mathrm{actor}}(\theta)
    =-\mathbb{E}\!\left[
      \frac1G\sum_{j=1}^{G}\frac1{L_j}\sum_{\ell=1}^{L_j}
      \min\!\left(
        \rho_{\theta}\widehat A_j,
        \operatorname{clip}(\rho_{\theta},1-\varepsilon,1+\varepsilon)\widehat A_j
      \right)
    \right]
    +\beta\,\operatorname{KL}(\pi_{\theta}\,\|\,\pi_{\mathrm{ref}}).
    \label{eq:grpo-actor-objective}
\end{equation}
\endgroup
Here, $\mathbb E$ averages over sampled tasks and rollout groups, $L_j$ counts
actor-generated tokens in trajectory $j$, and $\rho_{\theta}$ is the token-level
importance ratio between $\pi_\theta$ and $\pi_{\mathrm{old}}$.
The coefficient $\beta$ controls KL regularization toward the reference
policy $\pi_{\mathrm{ref}}$.

\subsection{Actor-Aligned Skill Proposal Learning}
\label{sec:proposal-learning}

\paragraph{Skill Proposal and Reference Construction.}
For a source trajectory $\tau\in\mathcal{T}_q$, we select
$\tau_{\mathrm{opposite}}$ from the opposite-outcome trajectories in the same
group and provide both trajectories to the skill proposer using
Eq.~\ref{eq:skill-proposal}.
For actor-alignment evaluation, we select one successful and one failed
reference trajectory, excluding both inputs $\tau$ and
$\tau_{\mathrm{opposite}}$ shown to the skill proposer.  Let $\mathcal{T}_q^+$
and $\mathcal{T}_q^-$ denote the successful and failed trajectories in the
group, respectively.  We sample
\begingroup
\setlength{\abovedisplayskip}{11pt plus2pt minus5pt}
\setlength{\abovedisplayshortskip}{6pt plus3pt}
\setlength{\belowdisplayskip}{8pt plus2pt minus5pt}
\setlength{\belowdisplayshortskip}{6pt plus3pt minus2pt}
\begin{equation}
    \tau^+\sim\mathcal{T}_q^+\setminus\{\tau,\tau_{\mathrm{opposite}}\},
    \qquad
    \tau^-\sim\mathcal{T}_q^-\setminus\{\tau,\tau_{\mathrm{opposite}}\}.
    \label{eq:input-excluded-references}
\end{equation}
\endgroup
Accordingly, we retain only trajectories for which both reference sets
remain nonempty after excluding the proposer inputs, with all trajectories
sharing the same task, retrieved skill $s$, and current actor.

\paragraph{Contrastive Action Feedback.}
We evaluate a candidate skill proposal by replacing the retrieved skill in the
reference action contexts.  We hold the pre-update policy, recorded histories, and action
tokens fixed; only the conditioning skill changes.
For a trajectory $\tau$, we average the token-normalized log-likelihoods of its
recorded actions under skill $s$ over the trajectory:
\vspace{-1pt}
\begin{equation}
    J(\tau;s)
    =\frac{1}{T}\sum_{t=1}^{T}
      \frac{1}{|a_t|}
      \log \pi_{\theta}\!\left(a_t\mid q,s,H_t\right),
    \label{eq:actor-relative-score}
\end{equation}
where $|a_t|$ is the token length of the recorded action.  The likelihood
changes induced by replacing $s$ with $\tilde{s}$ are
\begingroup
\setlength{\abovedisplayshortskip}{6pt plus3pt}
\setlength{\belowdisplayskip}{7pt plus2pt minus5pt}
\setlength{\belowdisplayshortskip}{6pt plus3pt minus2pt}
\begin{equation}
    \Delta^+ = J(\tau^+;\tilde{s})-J(\tau^+;s),
    \qquad
    \Delta^- = J(\tau^-;\tilde{s})-J(\tau^-;s),
\end{equation}
\endgroup
and the actor-alignment reward is
\begingroup
\setlength{\abovedisplayshortskip}{6pt plus3pt}
\setlength{\belowdisplayskip}{7pt plus2pt minus5pt}
\setlength{\belowdisplayshortskip}{6pt plus3pt minus2pt}
\begin{equation}
    R_{\mathrm{align}}
    = \Delta^+-\Delta^-.
    \label{eq:r-align}
\end{equation}
\endgroup
Thus, $R_{\mathrm{align}}>0$ indicates that the candidate increases the gap in
action log-likelihood between the successful and failed references relative
to the retrieved skill.

\paragraph{Skill Proposal Rewards.}
$R_{\mathrm{align}}$ evaluates behavioral alignment, but does not assess
proposal format, skill-edit operation appropriateness, or whether the proposed
skill content is grounded in the source trajectory.
The proposed skill content may also score well by reproducing
trajectory-specific details without yielding reusable guidance.

Accordingly, we check proposal format directly and prompt a skill critic to
judge whether the skill-edit operation is appropriate and the proposed content
is trajectory-grounded and reusable, yielding $d_{\mathrm{op}}$ and
$d_{\mathrm{skill}}$, respectively.  The critic supplies validity judgments
rather than a graded estimate of utility for the current actor.  For content
it deems supported, $R_{\mathrm{align}}$ provides actor-alignment feedback.
These signals yield the
proposal-format reward $r_{\mathrm{fmt}}$, the
skill-edit operation reward $r_{\mathrm{op}}$, and the skill-content reward
$r_{\mathrm{skill}}$.  Appendix~\ref{app:skill-critic-prompt} provides the
critic instruction.

Let $\eta>0$ be a fixed feedback magnitude.  Since format is checked first,
malformed proposals bypass the critic with $r_{\mathrm{op}}=r_{\mathrm{skill}}=0$;
otherwise, these two rewards are defined below.  The proposal-format and
skill-edit operation rewards are
\begin{equation}
    r_{\mathrm{fmt}}
    =
    \begin{cases}
        +\eta, & \text{if the proposal is well formed},\\
        -\eta, & \text{otherwise}
    \end{cases},
    \qquad
    r_{\mathrm{op}}
    =
    \begin{cases}
        +\eta, & d_{\mathrm{op}}=\textsc{Appropriate},\\
        -\eta, & d_{\mathrm{op}}=\textsc{Inappropriate}
    \end{cases}.
    \label{eq:proposal-structure-operation-reward}
\end{equation}
The skill-content reward is
\begin{equation}
    r_{\mathrm{skill}}
    =
    \begin{cases}
        \operatorname{clip}(R_{\mathrm{align}},-\eta,\eta),
        & \mathrm{op}\in\{\textsc{Add},\textsc{Update}\},\quad
          d_{\mathrm{skill}}=\textsc{Supported},\\
        -\eta,
        & \mathrm{op}\in\{\textsc{Add},\textsc{Update}\},\quad
          d_{\mathrm{skill}}=\textsc{Unsupported},\\
        0,
        & \mathrm{op}=\textsc{No Edit}
    \end{cases}.
    \label{eq:proposal-skill-reward}
\end{equation}
Clipping bounds the magnitude of $R_{\mathrm{align}}$ by $\eta$.
For well-formed proposals, the critic assesses edit appropriateness, checking
content support only for
\textsc{Add} and \textsc{Update}.  When content is supported, $R_{\mathrm{align}}$
provides actor-alignment feedback, whereas \textsc{No Edit} has neutral
$r_{\mathrm{skill}}=0$.

\paragraph{Proposer Optimization.}
Each eligible source trajectory yields one skill proposal.
Following prior work~\citep{zhang2026retroagent}, we optimize the proposer with
REINFORCE++~\citep{hu2025reinforcepp}.
We sum the format, skill-edit operation, and skill-content rewards into a scalar
proposal reward and normalize it across the global proposal batch:
\begin{equation}
    r_{\mathrm{prop}}
    =r_{\mathrm{fmt}}+r_{\mathrm{op}}+r_{\mathrm{skill}},
    \qquad
    \widehat A_{\mathrm{prop}}
    =\frac{r_{\mathrm{prop}}-\mu_{\mathrm{prop}}}
           {\sigma_{\mathrm{prop}}+\epsilon}.
    \label{eq:reinforcepp-proposal-advantage}
\end{equation}
Here, $\mu_{\mathrm{prop}}$ and $\sigma_{\mathrm{prop}}$ are the mean and standard
deviation of proposal rewards across the global batch, with each proposal
contributing one scalar reward.
We apply the same normalized advantage to all generated tokens in each skill
proposal and minimize the following loss:
\begin{equation}
    \mathcal{L}_{\mathrm{R++}}(\theta)
    =-\mathbb{E}\!\left[
      \mathbb E_t\!
      \min\!\left(
        \rho_{\theta}\widehat A_{\mathrm{prop}},
        \operatorname{clip}(\rho_{\theta},1-\varepsilon,1+\varepsilon)
        \widehat A_{\mathrm{prop}}
      \right)
    \right]
    +\beta_{\mathrm{prop}}\,
      \operatorname{KL}(\pi_\theta\,\|\,\pi_{\mathrm{ref}}).
    \label{eq:reinforcepp-proposal-objective}
\end{equation}
Here, $\mathbb E_t$ averages over generated tokens in each proposal, and the outer
$\mathbb E$ averages over the global proposal batch.  The token-level ratio
$\rho_\theta$ is evaluated at each token; $\beta_{\mathrm{prop}}$ weights KL.

\paragraph{Skill-Edit Support Regularization.}
The proposal-level advantage is applied to both the skill-edit operation and
the proposed content.  Consequently, a negative advantage can reduce the
probability of a skill-edit operation even when the skill critic judges it
appropriate.  Repeated suppression can make the operation rarely sampled,
limiting exploration as the actor and skillbank evolve.  We therefore introduce
\emph{Skill-Edit Support Regularization}, which penalizes available skill-edit
operations whose normalized probabilities fall below a threshold $p_{\min}$.
Let $\mathcal{O}(s)$ denote the available skill-edit operations.
\textsc{No~Edit} and \textsc{Add} are always available, while \textsc{Update}
is available only when a skill was retrieved for the source trajectory.
In the following equations, $x$ is shorthand for
$(q,s,\tau,\tau_{\mathrm{opposite}})$.
At the operation decision position, let $\pi_\theta(\mathrm{op}\mid x)$
denote the next-token probability of the first token of operation label
$\mathrm{op}$.
We then renormalize these probabilities over $\mathcal{O}(s)$:
\begin{equation}
    p_\theta(\mathrm{op}\mid x)
    =\frac{\pi_\theta(\mathrm{op}\mid x)}
    {\sum_{\mathrm{op}'\in\mathcal{O}(s)}
     \pi_\theta(\mathrm{op}'\mid x)},
    \qquad \mathrm{op}\in\mathcal{O}(s).
    \label{eq:operation-probability}
\end{equation}
The regularizer is
\begin{equation}
    \mathcal{L}_{\mathrm{sup}}(\theta)
    =\mathbb{E}_{x}\!\left[
      \frac{1}{|\mathcal{O}(s)|}
      \sum_{\mathrm{op}\in\mathcal{O}(s)}
      \left[
        \max\!\left(0,\log p_{\min}
        -\log p_\theta(\mathrm{op}\mid x)\right)
      \right]^2
    \right].
    \label{eq:operation-support-loss}
\end{equation}
The penalty is zero whenever all available operations have probabilities at
least $p_{\min}$.  It discourages premature suppression without requiring a
uniform skill-edit operation distribution.
Combining this regularizer with the REINFORCE++ loss yields
the complete skill proposer objective:
\begin{equation}
    \mathcal{L}_{\mathrm{proposer}}(\theta)
    =\mathcal{L}_{\mathrm{R++}}(\theta)
      +\lambda_{\mathrm{sup}}\mathcal{L}_{\mathrm{sup}}(\theta).
    \label{eq:reinforcepp-proposer-objective}
\end{equation}
The coefficient $\lambda_{\mathrm{sup}}$ controls the strength of support
regularization.

\subsection{Joint Policy--Skill Co-Evolution}
\label{sec:joint-evolution}

At each training step, actor rollouts and skill proposals are generated from
the same pre-update policy and skillbank snapshot.  We jointly optimize the
shared policy by minimizing
\begin{equation}
    \mathcal{L}_{\mathrm{UniSkill}}
    = \mathcal{L}_{\mathrm{actor}}
      +\lambda_{\mathrm{prop}}\mathcal{L}_{\mathrm{proposer}},
    \label{eq:joint-objective}
\end{equation}
where $\lambda_{\mathrm{prop}}$ weights the proposer loss.
A positive $R_{\mathrm{align}}$ indicates a wider action log-likelihood gap
between successful and failed references, even if likelihoods decrease on
both.  For skillbank updates, however, we adopt a stricter
criterion by additionally requiring $\Delta^+>0$ to avoid storing skills that
reduce the average action log-likelihood on the successful reference.
After the policy update, we apply \textsc{Add} and \textsc{Update} proposals
that pass both critic checks and satisfy $\Delta^+>0$ and
$R_{\mathrm{align}}>0$.  For each \textsc{Update} target, we apply the eligible proposal with the highest
$R_{\mathrm{align}}$.  The complete training procedure is summarized in Algorithm~\ref{alg:uniskill} in Appendix~\ref{app:training-algorithm}.

\section{Experiments}
\label{sec:experiments}

\subsection{Experimental Setup}

\paragraph{Benchmarks.}
We evaluate UniSkill on ALFWorld~\citep{shridhar2021alfworld} and
WebShop~\citep{yao2022webshop}, two multi-turn interactive benchmarks.
ALFWorld requires agents to complete household tasks through text-based
navigation and object manipulation.  We report per-task-type and overall
success rates.  WebShop requires an agent to search, inspect, and
purchase products that satisfy a natural-language instruction in a simulated
e-commerce website.  We report the mean normalized task score, scaled to 100,
and the success rate.

\paragraph{Baselines and Implementation Details.}
We compare UniSkill with training-free, RL-only, and memory- or
skill-augmented RL baselines.  All compared methods use
Qwen2.5-7B-Instruct~\citep{yang2024qwen25} as the base model.
Unless otherwise specified, the baseline results in
Table~\ref{tab:main-results} are drawn from
prior work~\citep{feng2025gigpo,shi2026skill1}, where they were obtained using the released
\textsc{verl-agent} implementations~\citep{feng2025gigpo} under a common
evaluation protocol.  We evaluate UniSkill using the same protocol, including
ALFWorld's official \texttt{valid\_seen} split, task-sampling procedure, and
success criterion.
Complementary RL results are quoted from its original
paper~\citep{muhtar2026complementary}, while
Evolving-RL~\citep{fan2026evolvingrl} is reproduced within
\textsc{verl-agent} under the ALFWorld training and evaluation setup used for
UniSkill.  UniSkill starts from an empty skillbank
and retrieves skills with Qwen3-Embedding-0.6B.  We prompt a DeepSeek-V4-Pro
model to serve as the skill critic.  We set $\eta=0.05$ and
$p_{\min}=0.1$.  Appendix~\ref{app:implementation-details} explains these
choices and provides detailed \mbox{training and generation configurations}.

\subsection{Performance on ALFWorld and WebShop}

As shown in Table~\ref{tab:main-results}, UniSkill achieves strong performance
on both benchmarks, reaching 98.4\% success on ALFWorld and a task score of
90.5 with 84.7\% success on WebShop.  Compared with RL-only baselines, UniSkill
improves ALFWorld success over GRPO and GiGPO by 20.8 and 7.6 percentage points (pp),
respectively.  Its advantage extends to skill-augmented RL, with a
12.0~pp improvement in success rate over SkillRL on WebShop.  UniSkill achieves
0.9~pp higher ALFWorld success than Skill1 while delivering a comparable task
score and 1.8~pp higher success on WebShop.

Evolving-RL provides a closely related joint-training baseline that evaluates
proposed skills through additional environment rollouts.  UniSkill exceeds
Evolving-RL by 5.4~pp in ALFWorld success.
Beyond final performance, Figure~\ref{fig:alfworld-learning-dynamics}(a)
reveals contrasting training dynamics.  Evolving-RL makes strong early gains
but subsequently undergoes performance collapse, a risk also examined in their
original study~\citep{fan2026evolvingrl}.  In contrast, UniSkill sustains stable
joint training over 250 steps, with validation success showing an overall
\mbox{upward trend and continued improvement during later training.}
Alongside this performance trend, Figure~\ref{fig:alfworld-learning-dynamics}(b)
shows rapid growth of the learned skillbank early in training, followed by
slower expansion as accepted \textsc{Add} operations become less frequent,
while occasional additions and updates continue throughout training.

To further assess training efficiency,
\mbox{Figure~\ref{fig:training-efficiency}} compares the per-step time
spent on actor rollouts, proposal generation, skill evaluation, and skill
critic calls.  Rollout-based skill evaluation accounts for most of
Evolving-RL's measured time, whereas UniSkill evaluates proposed skills using
$R_{\mathrm{align}}$ computed from previously collected trajectories.  The
resulting savings outweigh UniSkill's longer actor rollouts and skill critic
calls, yielding lower total time across the reported stages.
\looseness=-1

\begingroup
\setlength{\intextsep}{5pt}
\begin{figure}[!b]
    \centering
    \setlength{\abovecaptionskip}{4pt}
    \begin{minipage}[t]{0.47\linewidth}
        \centering
        \includegraphics[width=0.92\linewidth]{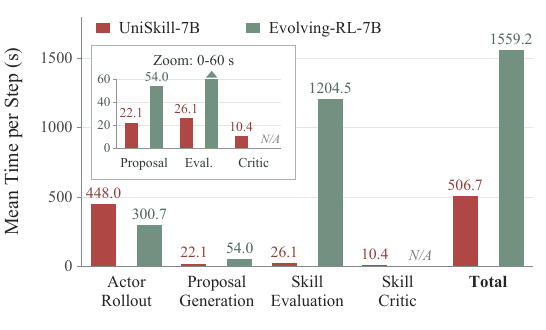}
        \caption{Measured per-step time for UniSkill-7B and Evolving-RL-7B on ALFWorld.}
        \label{fig:training-efficiency}
    \end{minipage}\hfill
    \begin{minipage}[t]{0.47\linewidth}
        \centering
        \includegraphics[width=0.92\linewidth]{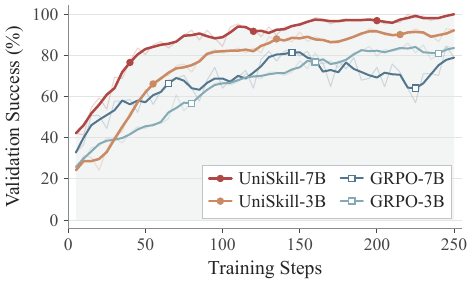}
        \caption{ALFWorld validation at 3B and 7B; 3-step moving averages over faint raw curves.}
        \label{fig:model-size-comparison}
    \end{minipage}
\end{figure}
\endgroup

\begin{table}[!t]
    \centering
    \small
    \setlength{\belowcaptionskip}{6pt}
    \caption{Results on ALFWorld and WebShop.  We report success rates
    (\%) and normalized WebShop scores.  UniSkill reports mean $\pm$ std
    over three independent training runs.
    Best and second-best results are bold and underlined.
    $^{\dagger}$ denotes purchase-completion rate, which is excluded from ranking.}
    \setlength{\tabcolsep}{2.6pt}
    \newcommand{\centeredmeanstd}[2]{%
        \begingroup
        \setbox0=\hbox{#1$_{\pm #2}$}%
        \makebox[\wd0][c]{#1\rlap{$_{\pm #2}$}}%
        \endgroup
    }
    \newsavebox{\mainresultstablebox}
    \savebox{\mainresultstablebox}{\resizebox{\linewidth}{!}{
    \begin{tabular}{lccccccc@{\hspace{10.6pt}}|cc@{\hspace{10.6pt}}}
        \toprule
        \multirow{2}{*}{\textbf{Method}} & \multicolumn{7}{c|}{\textbf{ALFWorld (Success \%)}}
        & \multicolumn{2}{c}{\textbf{WebShop}} \\
        \cmidrule(lr){2-8}\cmidrule(lr){9-10}
        & Pick & Look & Clean & Heat & Cool & Pick2 & All
        & Score & Succ. \\
        \midrule
        \rowcolor[gray]{0.92}\multicolumn{10}{c}{\textit{Training-Free Methods}} \\
        Base Model & 33.4 & 21.6 & 19.3 & 6.9 & 2.8 & 3.2 & 14.8 & 26.4 & 7.8 \\
        ReAct~\citep{yao2023react} & 48.5 & 35.4 & 34.3 & 13.2 & 18.2 & 17.6 & 31.2 & 46.2 & 19.5 \\
        Reflexion~\citep{shinn2023reflexion} & 62.0 & 41.6 & 44.9 & 30.9 & 36.3 & 23.8 & 42.7 & 58.1 & 28.8 \\
        ExpeL~\citep{zhao2024expel} & 21.0 & 67.0 & 55.0 & 52.0 & 71.0 & 6.0 & 46.3 & 30.9 & 11.2 \\
        \midrule
        \rowcolor[gray]{0.92}\multicolumn{10}{c}{\textit{Reinforcement Learning Methods}} \\
        PPO~\citep{schulman2017ppo} & 92.3 & 64.0 & 92.5 & 89.5 & 80.3 & 68.8 & 80.4 & 81.4 & 68.7 \\
        RLOO~\citep{ahmadian2024rloo} & 87.6 & 78.2 & 87.3 & 81.3 & 71.9 & 48.9 & 75.5 & 80.3 & 65.7 \\
        GRPO~\citep{shao2024deepseekmath} & 90.8 & 66.1 & 89.3 & 74.7 & 72.5 & 64.7 & 77.6 & 79.3 & 66.1 \\
        GiGPO~\citep{feng2025gigpo} & 97.7 & 82.7 & 98.8 & 83.7 & 89.3 & 79.2 & 90.8 & 84.4 & 72.8 \\
        \midrule
        \rowcolor[gray]{0.92}\multicolumn{10}{c}{\textit{Memory- or Skill-Augmented Reinforcement Learning Methods}} \\
        EvolveR~\citep{wu2026evolver} & 64.9 & 33.3 & 46.4 & 13.3 & 33.3 & 33.3 & 43.8 & 42.5 & 17.6 \\
        Mem0 \textit{(w/ GRPO)}~\citep{chhikara2025mem0} & 78.1 & 54.8 & 56.1 & 31.0 & 65.0 & 26.9 & 54.7 & 58.1 & 37.5 \\
        SimpleMem \textit{(w/ GRPO)}~\citep{liu2026simplemem} & 89.5 & 36.3 & 60.0 & 50.0 & 64.9 & 26.3 & 62.5 & 67.8 & 46.9 \\
        SkillRL~\citep{xia2026skillrl} & \underline{97.9} & 71.4 & 90.0 & 90.0 & 95.5 & 87.5 & 89.9 & 85.2 & 72.7 \\
        Complementary RL~\citep{muhtar2026complementary} & -- & -- & -- & -- & -- & -- & 91.0 & -- & 87.0\rlap{$^{\dagger}$} \\
        RetroAgent~\citep{zhang2026retroagent} & \underline{97.9} & 90.9 & \underline{99.2} & 92.9 & 85.3 & 91.0 & 94.9 & 88.9 & 82.3 \\
        Skill1~\citep{shi2026skill1} & \textbf{100.0} & \underline{98.6} & 97.3 & \textbf{99.2} & \underline{96.1} & \underline{96.0} & \underline{97.5} & \underline{89.7} & \underline{82.9} \\
        Evolving-RL~\citep{fan2026evolvingrl} & 97.1 & \textbf{100.0} & 92.6 & 87.5 & 88.0 & 91.7 & 93.0 & -- & -- \\
        \midrule
        \textbf{UniSkill (Ours)} & \centeredmeanstd{\textbf{100.0}}{0.0} & \centeredmeanstd{\textbf{100.0}}{0.0} & \centeredmeanstd{\textbf{100.0}}{0.0} & \centeredmeanstd{\underline{93.0}}{6.1} & \centeredmeanstd{\textbf{100.0}}{0.0} & \centeredmeanstd{\textbf{96.7}}{2.9} & \centeredmeanstd{\textbf{98.4}}{0.8} & \centeredmeanstd{\textbf{90.5}}{1.4} & \centeredmeanstd{\textbf{84.7}}{0.5} \\
        \bottomrule
    \end{tabular}}}
    \makebox[0pt][l]{%
        \hspace{\tabcolsep}%
        \smash{\raisebox{\dimexpr-\dp\mainresultstablebox+0.7pt\relax}{%
            {\color{blue!6}\rule{\dimexpr\wd\mainresultstablebox-\tabcolsep\relax}{2.2ex}}}}}%
    \usebox{\mainresultstablebox}
    \label{tab:main-results}
\end{table}

\subsection{Performance Across Model Sizes}
\label{sec:model-sizes}

We examine whether UniSkill remains effective with Qwen2.5-3B-Instruct as the
shared actor and skill proposer backbone, comparing it with size-matched GRPO
references in \mbox{Figure~\ref{fig:model-size-comparison}}.  Validation success
trends upward at both sizes, with UniSkill-3B continuing to improve during
later training, exceeding 90\% success, and finishing above \mbox{GRPO-7B}.
Together, these results suggest that
policy--skill co-evolution remains
effective with a smaller shared backbone.
\looseness=-1

\subsection{Additional Experiments}
\label{sec:additional-experiments}

\noindent\textbf{(1) Out-of-distribution generalization.}
UniSkill performs well on the ALFWorld out-of-distribution split with and
without skill retrieval, with retrieval providing further gains, as detailed
in Appendix~\ref{app:ood-generalization}.
\textbf{(2) Sensitivity to the skill critic.}
UniSkill achieves comparable WebShop performance with different skill critic
models, as detailed in Appendix~\ref{app:critic-sensitivity}.

\Needspace{4\baselineskip}
\section{Analysis}
\label{sec:analysis}

In this section, we present an empirical analysis of UniSkill by addressing
the following questions:

\paragraph{Q1: How Does Joint Actor and Skill Proposer Learning Contribute to Performance?}
To assess the contribution of joint learning, we compare UniSkill with three
ablation variants in Table~\ref{tab:joint-learning-ablation}.
\emph{\mbox{Actor Only}} trains the actor with a frozen skill proposer, whereas
\emph{Skill Proposer Only} trains the proposer with a frozen actor.  In both
variants, the frozen model retains the initial Qwen2.5-7B-Instruct weights.
\emph{UniSkill (w/o $R_{\mathrm{align}}$ reward)} removes $R_{\mathrm{align}}$
only from the proposer reward.  All variants start from an empty skillbank
and retain skill retrieval during training, with identical critic checks and
skillbank update rules.

\begingroup
\setlength{\intextsep}{2pt}
\begin{table}[H]
    \centering
    \setlength{\belowcaptionskip}{4pt}
    \caption{Ablation results on ALFWorld with and without skill retrieval at evaluation.}
    \label{tab:joint-learning-ablation}
    \small
    \setlength{\tabcolsep}{5pt}
    \renewcommand{\arraystretch}{1.0}
    \begin{tabular*}{\linewidth}{@{\extracolsep{\fill}}lccccc@{}}
        \toprule
        \multirow{2}{*}{Setting}
        & \multicolumn{3}{c}{Training}
        & \multicolumn{2}{c}{ALFWorld Success Rate (\%)} \\
        \cmidrule(lr){2-4}\cmidrule(lr){5-6}
        & Actor
        & Skill Proposer
        & $R_{\mathrm{align}}$ reward
        & w/o Retrieval
        & w/ Retrieval \\
        \midrule
        Actor Only
        & Updated & Frozen & $\times$ & 84.4 & 87.5 \\
        Skill Proposer Only
        & Frozen & Updated & $\checkmark$ & 34.4 & 32.8 \\
        UniSkill (w/o $R_{\mathrm{align}}$ reward)
        & Updated & Updated & $\times$ & 84.4 & 89.1 \\
        UniSkill
        & Updated & Updated & $\checkmark$ & 96.1 & 98.4 \\
        \bottomrule
    \end{tabular*}
\end{table}
\endgroup

Table~\ref{tab:joint-learning-ablation} shows that UniSkill retains 96.1\%
success when skill retrieval is disabled at evaluation, compared with 84.4\%
for both Actor Only and the variant without the $R_{\mathrm{align}}$ reward,
and 34.4\% for Skill Proposer Only.  Since the variant without the
$R_{\mathrm{align}}$ reward follows the same joint-training procedure and
skillbank update rules, differing only in whether $R_{\mathrm{align}}$ is included
in the training objective of the proposer, this comparison identifies the contribution
of actor-alignment feedback as a training signal to the actor's no-retrieval
performance.  By evaluating self-proposed skills for alignment with the current actor,
this feedback favors guidance that can shape subsequent skill-conditioned
learning, consistent with the continued rise in validation success during
later training and the strong performance of the actor without retrieval.  When
evaluated with its evolved skillbank, UniSkill reaches 98.4\% success,
outperforming all three variants and providing an additional benefit alongside
the gains retained by the actor.
\looseness=-1

\paragraph{Q2: Does $R_{\mathrm{align}}$ Capture Actor Alignment?}
The ablation results in Q1 show that $R_{\mathrm{align}}$ benefits proposer
training.  We next examine whether higher scores indicate better
\emph{actor alignment}, measured by the success-rate gain from replacing the
retrieved skill with the proposed skill under the same actor.  Specifically, at
checkpoint $k$, we compute $R_{\mathrm{align}}$ and evaluate each source task
$q_i$ with the fixed actor $\pi^{(k)}$, directly supplying either the retrieved
or proposed skill.
The rollout gain is
\begin{equation}
    U_i^{(k)}
    = \mathbb{E}\!\left[R \mid q_i,\tilde{s}_i,\pi^{(k)}\right]
      - \mathbb{E}\!\left[R \mid q_i,s_i,\pi^{(k)}\right],
    \label{eq:empirical-actor-alignment}
\end{equation}
where $R$ denotes the task return.  We estimate $U_i^{(k)}$ using mean task
returns from 32 rollouts per skill condition.
We evaluate critic-accepted \textsc{Add} and \textsc{Update} proposals with
clipped $R_{\mathrm{align}}$ at training steps 25 and 75; the sampling
procedure is detailed in Appendix~\ref{app:actor-alignment-evaluation}.  We
report Spearman correlations and linear trends with pointwise 95\% bootstrap
confidence intervals.
Figure~\ref{fig:alignment-validation} shows positive Spearman correlations
between $R_{\mathrm{align}}$ and success-rate gains at steps 25 and 75
($\rho=0.312$ and $0.367$, respectively).  Proposals with positive
$R_{\mathrm{align}}$ have mean gains of $8.4$ and $7.2$ percentage points,
compared with $-3.4$ and $-0.5$ percentage points for negative-score proposals.
However, $13/53$ (24.5\%) and $10/64$ (15.6\%) of positive-score proposals
have negative observed gains at the two checkpoints, respectively.  Thus,
$R_{\mathrm{align}}$ is informative in aggregate among the sampled proposals
but does not guarantee improvement for every proposal.
\looseness=-1

\begingroup
\setlength{\intextsep}{4pt}
\begin{figure}[!htbp]
    \centering
    \setlength{\abovecaptionskip}{0pt}
    \includegraphics[width=0.78\linewidth]{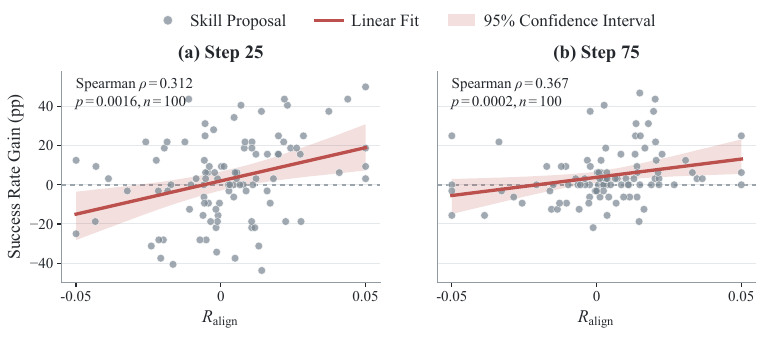}
    \caption{Actor alignment on ALFWorld with linear fits and pointwise 95\%
    bootstrap CIs.}
    \label{fig:alignment-validation}
\end{figure}
\endgroup

\Needspace{11\baselineskip}
\paragraph{Q3: Does Support Regularization Sustain Skill-Edit Exploration?}
The analysis in Q2 supports $R_{\mathrm{align}}$ as a proxy for actor alignment.
As discussed in Section~\ref{sec:proposal-learning}, however, this feedback
alone does not ensure sustained skill-edit exploration.  We therefore compare UniSkill with and without
support regularization, examining skill-edit distributions and validation
performance.
\looseness=-1

\begingroup
\setlength{\intextsep}{4pt}
\begin{figure}[!htbp]
    \centering
    \setlength{\abovecaptionskip}{0pt}
    \includegraphics[width=\linewidth]{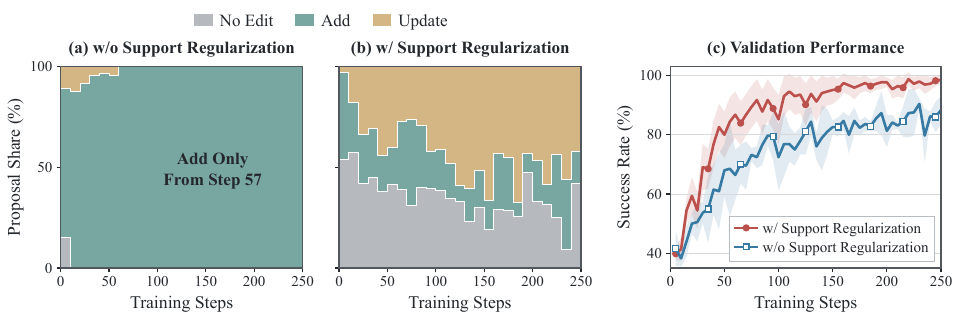}
    \caption{Support regularization on ALFWorld.
    (a--b) Skill-edit distributions (one run per setting; 10-step windows).
    (c) Validation success (mean $\pm$ std; three independent runs per setting).}
    \label{fig:skill-edit-exploration}
    \vspace{6pt}
\end{figure}
\endgroup

As shown in Figure~\ref{fig:skill-edit-exploration}(a), the run without support
regularization exhibits a collapse to a single skill-edit operation
(\textsc{Add} in this run; see
Appendix~\ref{app:additional-skill-edit-dynamics} for additional cases).
In contrast, Figure~\ref{fig:skill-edit-exploration}(b) shows that the run
with support regularization continues to sample all three operations with
changing proportions, consistent with preserving exploration rather than
enforcing a uniform distribution.  The validation curves in
Figure~\ref{fig:skill-edit-exploration}(c) further show faster early improvement
and higher success rates during later training with support regularization.
Preserving these exploration opportunities allows the skill proposer to adapt
its edit choices as the actor and skillbank evolve, supporting stable joint
training.
\looseness=-1

\section{Conclusion}

We present UniSkill, which jointly learns task execution and skill proposals
with a shared policy, enabling the actor and skillbank to co-evolve.
Actor-aligned feedback from previously collected trajectories reduces
skill-evaluation overhead, while support regularization sustains exploration
across skill-edit operations throughout training.  Experiments on ALFWorld and
WebShop demonstrate strong task performance and stable joint training.  Further
ALFWorld experiments show that UniSkill remains effective when the shared
policy uses a smaller backbone.
\looseness=-1

\section*{Ethics Statement}

This work evaluates UniSkill in the publicly available ALFWorld and WebShop
benchmark environments, without human-subject experiments or collection of
personal data.  The experiments do not involve real-world physical actions
or purchases.
\section*{Reproducibility Statement}

Section~\ref{sec:methodology} specifies the training objectives, contrastive action feedback,
and skillbank update rules.  Algorithm~\ref{alg:uniskill} in
Appendix~\ref{app:training-algorithm} summarizes the training procedure.
Section~\ref{sec:experiments} describes the benchmarks, evaluation metrics, and baseline sources.
Appendix~\ref{app:prompts} provides the actor, skill proposer, and skill critic
prompts.  Appendix~\ref{app:implementation-details} reports training and
generation configurations, the ALFWorld evaluation protocol, and the choices
of $\eta$ and $p_{\min}$.  Section~\ref{sec:analysis} describes the ablations and statistical
analyses, with query selection and proposal sampling detailed in
Appendix~\ref{app:actor-alignment-evaluation}.  We plan to publicly release
the implementation and configuration files.

\section*{AI Use Statement}

The authors led the research, implementation, experimental execution, and
manuscript preparation.  Generative AI tools provided assistance with
literature review, methodological and experimental discussions, translation,
editing, and \LaTeX{} formatting.  The authors take full responsibility for
this paper.

\bibliography{iclr2027_conference}
\bibliographystyle{iclr2027_conference}

\clearpage
\appendix
\section{Prompts}
\label{app:prompts}

\tcbset{uniskillprompt/.style={
    enhanced, breakable, width=\linewidth, center,
    colback=white, colframe=black!25, colbacktitle=black!5, coltitle=black,
    boxrule=0.45pt, arc=0.7mm,
    left=3mm, right=3mm, top=2.5mm, bottom=2.5mm,
    before skip=10pt, after skip=8pt, fonttitle=\small\bfseries,
    before upper={\small\raggedright\setlength{\parindent}{0pt}\setlength{\parskip}{4pt}}
}}

\subsection{ALFWorld Prompts}
\label{app:alfworld-prompts}

\begin{tcolorbox}[uniskillprompt, title={ALFWorld Actor Prompt}]
You are in ACTION MODE, acting in the ALFRED Embodied Environment.

\textbf{Task:} \texttt{\{task\}}\\
\textbf{Retrieved skill:} \texttt{\{skill\}}

\textbf{Recent observations and actions:} \texttt{\{history\}}\\
\textbf{Current step:} \texttt{\{step\}}\\
\textbf{Current observation:} \texttt{\{observation\}}\\
\textbf{Admissible actions:} \texttt{\{admissible\_actions\}}

First reason step-by-step within \texttt{<think>...</think>}.  Then output one
admissible action for the current step within \texttt{<action>...</action>}.
\end{tcolorbox}

\begin{tcolorbox}[uniskillprompt, title={ALFWorld Skill Proposer Prompt}]
You are in SKILL PROPOSAL MODE.  Compare the completed ALFWorld episode with
the opposite-outcome trajectory and propose one skill-edit operation.

\textbf{Task type:} \texttt{\{task\_type\}}\\
\textbf{Task:} \texttt{\{task\}}\\
\textbf{Retrieved skill:} \texttt{\{retrieved\_skill\}}

\textbf{Current episode.}\\
Outcome: \texttt{\{success\_and\_end\_reason\}}\\
Trajectory: \texttt{\{observations, reasoning, and actions\}}

\textbf{Opposite-outcome trajectory.}\\
Outcome: \texttt{\{reference\_success\_and\_end\_reason\}}\\
Trajectory: \texttt{\{reference\_actions\}}

Use the comparison to identify reusable behavioral differences supported by
the recorded actions and outcomes.

\tcbline
\textbf{Choose one available operation.}
\begin{description}
    \setlength{\itemsep}{2pt}
    \setlength{\parskip}{0pt}
    \item[\texttt{NO\_EDIT}] Choose this if the episode does not reveal a
    reusable skill.
    \item[\texttt{ADD\_NEW\_SKILL}] Choose this if the episode reveals a new
    reusable skill.
    \item[\texttt{UPDATE\_SKILL}] Choose this if the retrieved skill should
    be improved based on the episode.  This operation is available only when
    a skill was retrieved.
\end{description}

A skill must be concrete, supported by the source trajectory, and usable from the
actor's available prompt.  Do not merely restate the task, copy the retrieved
skill, or memorize this episode.  For \texttt{ADD\_NEW\_SKILL} or
\texttt{UPDATE\_SKILL}, write one self-contained reusable skill in plain
text.  For \texttt{NO\_EDIT}, write exactly \texttt{NONE} in the skill block.

\tcbline
\textbf{Response format.}
Return exactly two tagged blocks:

\texttt{<action>\{operation\}</action>}\\
\texttt{<skill>\{skill text or NONE\}</skill>}
\end{tcolorbox}

\subsection{WebShop Prompts}
\label{app:webshop-prompts}

\begin{tcolorbox}[uniskillprompt, title={WebShop Actor Prompt}]
You are an expert autonomous agent operating in the WebShop e-commerce
environment.

\textbf{Retrieved skill:} \texttt{\{skill\}}\\
\textbf{Task:} \texttt{\{task\}}\\
\textbf{Recent observations and actions:} \texttt{\{history\}}\\
\textbf{Current step:} \texttt{\{step\}}\\
\textbf{Current observation:} \texttt{\{observation\}}\\
\textbf{Admissible actions:} \texttt{\{admissible\_actions\}}

Reason about the shopping goal within \texttt{<think>...</think>}.  Then output
one admissible action within \texttt{<action>...</action>}.
\end{tcolorbox}

\begin{tcolorbox}[uniskillprompt, title={WebShop Skill Proposer Prompt}]
You are in SKILL PROPOSAL MODE.  Compare the completed WebShop episode with
the opposite-outcome trajectory and propose one skill-edit operation.

\textbf{Task:} \texttt{\{task\}}\\
\textbf{Retrieved skill:} \texttt{\{retrieved\_skill\}}

\textbf{Current episode.}\\
Outcome: \texttt{\{success, task score, and end reason\}}\\
Action sequence: \texttt{\{source\_actions\}}

\textbf{Opposite-outcome trajectory.}\\
Outcome: \texttt{\{success, task score, and end reason\}}\\
Action sequence: \texttt{\{reference\_actions\}}

Use the comparison to identify reusable behavioral differences supported by
the recorded actions and outcomes.  Do not infer hidden causes, memorize
product identifiers, or copy an exact action path.

\tcbline
\textbf{Choose one available operation.}
\begin{description}
    \setlength{\itemsep}{2pt}
    \setlength{\parskip}{0pt}
    \item[\texttt{NO\_EDIT}] No concrete reusable guidance is revealed, or
    the retrieved skill already contains it.
    \item[\texttt{ADD\_NEW\_SKILL}] Distinct reusable guidance should be
    stored as a separate skill.
    \item[\texttt{UPDATE\_SKILL}] Revise an existing retrieved skill based
    on the comparison.
\end{description}

Write concrete, self-contained guidance reusable across shopping tasks.
Do not restate the request or copy or paraphrase existing skills.  Use
\texttt{NONE} for \texttt{NO\_EDIT}.

\tcbline
\textbf{Response format.}
Return exactly two tagged blocks:

\texttt{<action>\{operation\}</action>}\\
\texttt{<skill>\{skill text or NONE\}</skill>}
\end{tcolorbox}

\subsection{Skill Critic Prompt}
\label{app:skill-critic-prompt}

\begin{tcolorbox}[
    enhanced,
    breakable,
    width=\linewidth,
    center,
    colback=white,
    colframe=black!25,
    colbacktitle=black!5,
    coltitle=black,
    boxrule=0.45pt,
    arc=0.7mm,
    left=3mm,
    right=3mm,
    top=2.5mm,
    bottom=2.5mm,
    before skip=10pt,
    after skip=8pt,
    title={\textbf{Skill Critic Prompt}},
    fonttitle=\small
]
\small
\raggedright
\setlength{\parindent}{0pt}

\textbf{Task and inputs.}
You are a skill critic.  Independently assess whether a proposed skill-edit
operation is reasonable and whether its content passes a permissive validity
check.  You receive:

\begin{description}
    \setlength{\itemsep}{1pt}
    \setlength{\parskip}{0pt}
    \item[\texttt{TRAJECTORY}] The task, observations, actions, and
    environment-confirmed outcome.
    \item[\texttt{RETRIEVED\_SKILL}] The retrieved skill, or \texttt{NONE}.
    \item[\texttt{PROPOSED\_ACTION}] \texttt{NO\_EDIT},
    \texttt{ADD\_NEW\_SKILL}, or \texttt{UPDATE\_SKILL}.
    \item[\texttt{PROPOSED\_SKILL}] The candidate skill, or \texttt{NONE}
    for \texttt{NO\_EDIT}.
\end{description}

\tcbline
\textbf{1. Assess the skill-edit operation.}
Determine whether the trajectory provides concrete reusable guidance beyond
the retrieved skill.  Successful trajectories may demonstrate decision rules
or procedures; failures may reveal corrections to omissions, action ordering,
or repeated mistakes.  A correction need not have been executed successfully,
and standard procedures may still qualify.  Restating the task does not.
Judge the proposed operation using the following rules:

\begin{description}
    \setlength{\itemsep}{1pt}
    \setlength{\parskip}{0pt}
    \item[\texttt{NO\_EDIT}] No new reusable guidance is present, or the
    retrieved skill already contains the required guidance and the trajectory
    merely failed to follow it.
    \item[\texttt{ADD\_NEW\_SKILL}] The guidance should stand as a separate
    reusable skill, including when \texttt{RETRIEVED\_SKILL} is \texttt{NONE}.
    \item[\texttt{UPDATE\_SKILL}] The guidance should revise or extend the
    retrieved skill; an existing skill is required.
\end{description}

When adding and updating are genuinely ambiguous, accept either.
Ignore candidate wording, quality, support, usefulness, specificity, and
effectiveness when judging the operation.

\tcbline
\textbf{2. Assess the skill content.}
For \texttt{ADD\_NEW\_SKILL} and \texttt{UPDATE\_SKILL}, check validity
independently of the operation verdict.  Do not rank skill quality or require
proven effectiveness.

\textbf{Accept} content that is trajectory-grounded, reusable, actionable,
and plausibly useful.  Imperfect, incomplete, simple, standard, or suboptimal
guidance may pass.  An observed mistake can support a correction that was not
executed.

\textbf{Reject only} content that is clearly unsupported or contradictory;
irrelevant or too vague to guide behavior; repeats a pattern that clearly
contributed to failure; merely restates the task or outcome; copies
instance-specific identifiers, locations, or an exact trajectory without
reusable abstraction; or contains noise or manipulation.

\tcbline
\textbf{3. Return the verdict.}
Return only a JSON object with these four fields:

\begin{description}
    \setlength{\itemsep}{1pt}
    \setlength{\parskip}{0pt}
    \item[\texttt{action\_reasonable}] \texttt{true} or \texttt{false}.
    \item[\texttt{action\_reason}] A brief reason for the skill-edit operation.
    \item[\texttt{content\_supported}] \texttt{true} or \texttt{false} for
    \texttt{ADD\_NEW\_SKILL} and \texttt{UPDATE\_SKILL}; \texttt{null} for
    \texttt{NO\_EDIT}.
    \item[\texttt{content\_reason}] A brief reason for the proposed content.
\end{description}

\end{tcolorbox}

\section{Implementation Details}
\label{app:implementation-details}

UniSkill is implemented on \textsc{verl} and \textsc{verl-agent}.
Unless otherwise specified, the actor and skill proposer share a
Qwen2.5-7B-Instruct policy and an AdamW optimizer.  We accumulate gradients
from both objectives before each optimizer update.
Table~\ref{tab:training-configuration} lists the shared hyperparameters and
benchmark-specific settings.  We use a frozen DeepSeek-V4-Pro skill critic
and retrieve the top-1 skill using Qwen3-Embedding-0.6B.  Each run starts
from an empty skillbank.

\begin{table}[!htbp]
    \centering
    \small
    \caption{Training and generation settings.}
    \label{tab:training-configuration}
    \setlength{\tabcolsep}{6pt}
    \renewcommand{\arraystretch}{1.02}
    \begin{tabular*}{\linewidth}{@{\extracolsep{\fill}}lcc@{}}
        \toprule
        \textbf{Parameter} & \textbf{ALFWorld} & \textbf{WebShop} \\
        \midrule
        \rowcolor[gray]{0.92}\multicolumn{3}{c}{\textit{Optimization}} \\
        Optimizer & AdamW & AdamW \\
        AdamW $(\beta_1,\beta_2)$ & $(0.9,0.999)$ & $(0.9,0.999)$ \\
        Learning rate (constant) & $1\times10^{-6}$ & $1\times10^{-6}$ \\
        Weight decay & 0.01 & 0.01 \\
        Gradient norm clipping & 1.0 & 1.0 \\
        Policy ratio clipping $\varepsilon$ & 0.2 & 0.2 \\
        KL weights $(\beta,\beta_{\mathrm{prop}})$ & $(0.01,0.02)$ & $(0.01,0.02)$ \\
        Loss weights $(\lambda_{\mathrm{prop}},\lambda_{\mathrm{sup}})$ & $(0.5,0.01)$ & $(0.5,0.01)$ \\
        Format-only proposer warm-up (steps) & 3 & 3 \\
        Feedback magnitude $\eta$ & 0.05 & 0.05 \\
        Operation probability floor $p_{\min}$ & 0.1 & 0.1 \\
        Update epochs & 1 & 1 \\
        \midrule
        \rowcolor[gray]{0.92}\multicolumn{3}{c}{\textit{Sampling and Generation}} \\
        Tasks per training step & 16 & 16 \\
        Rollouts per task $G$ & 8 & 8 \\
        Actor minibatch size (global responses) & 128 & 64 \\
        Microbatch size per GPU & 8 & 8 \\
        Training steps & 250 & 250 \\
        Training temperature (actor and skill proposer) & 1.0 & 1.0 \\
        Evaluation temperature & 0.4 & 0.4 \\
        Top-$p$ & 1.0 & 1.0 \\
        Maximum input length (tokens) & 16,384 & 16,384 \\
        Maximum actor response length (tokens) & 512 & 512 \\
        Maximum skill proposal length (tokens) & 1,024 & 1,024 \\
        Maximum environment interaction steps & 50 & 15 \\
        \bottomrule
    \end{tabular*}
\end{table}

Actor minibatch sizes count single-turn responses across all GPUs; the actor
output limit applies to each turn.  One skill proposal is generated per
selected source trajectory, so the global proposal batch size varies by
training step.

\paragraph{ALFWorld Evaluation.}
\label{app:alfworld-evaluation}
We follow the \textsc{verl-agent} evaluation pipeline, retaining its
task-sampling procedure and success criterion.  We use ALFWorld's
official \texttt{valid\_seen} split for the main results and
\texttt{valid\_unseen} split for out-of-distribution evaluation.
For the main and out-of-distribution results, we report the mean
and sample standard deviation over three independent training runs.  Each run
is evaluated using its own trained policy and associated skillbank, which
remain fixed during evaluation.  Overall success is averaged over all
episodes in each evaluation.

\paragraph{Evolving-RL Training Stability.}
The collapse observed in our Evolving-RL reproduction during extended
training (Figure~\ref{fig:alfworld-learning-dynamics}) is consistent with
the authors' public clarification that training beyond 100 steps becomes
increasingly unstable and eventually collapses.\footnote{Additional discussion
of extended training is available in the
\href{https://github.com/Fanzy27/Evolving-RL}{Evolving-RL repository}.}

\paragraph{Choice of $\eta$.}
We set $\eta=0.05$ based on the distribution of $R_{\mathrm{align}}$ in a
preliminary ALFWorld run.  As shown in Table~\ref{tab:eta-selection}, this
threshold preserves most raw scores while limiting the magnitude of larger
positive and negative feedback.

\begin{table}[H]
    \small
    \caption{Unclipped $R_{\mathrm{align}}$ statistics from a preliminary
    ALFWorld run ($n=566$).}
    \label{tab:eta-selection}
    \begin{center}
    \setlength{\tabcolsep}{8pt}
    \begin{tabular}{cccccc}
        \toprule
        \multicolumn{3}{c}{Percentiles of $|R_{\mathrm{align}}|$}
        & \multicolumn{3}{c}{Score proportions (\%)} \\
        \cmidrule(lr){1-3}\cmidrule(lr){4-6}
        50th & 90th & 95th
        & $R_{\mathrm{align}}<-\eta$
        & $|R_{\mathrm{align}}|\leq\eta$
        & $R_{\mathrm{align}}>\eta$ \\
        \midrule
        0.0060 & 0.0358 & 0.0615 & 3.0 & 93.3 & 3.7 \\
        \bottomrule
    \end{tabular}
    \end{center}
\end{table}

\paragraph{Choice of $p_{\min}$.}
The threshold is intended to preserve exploration opportunities rather than
enforce a uniform distribution over skill-edit operations.  We use the actor
rollout group size $G=8$ as a reference.  In a simplified model of independent
draws, let $p$ be the selection probability of a given available operation and
$N_{\mathrm{op}}$ its count.  Taking a minimum 50\% chance of at least one
occurrence as a heuristic reference gives
\begin{equation}
    \begin{gathered}
        \Pr(N_{\mathrm{op}}=0)=(1-p)^G,
        \qquad \Pr(N_{\mathrm{op}}\geq 1)=1-(1-p)^G, \\
        \Pr(N_{\mathrm{op}}\geq 1)\geq 0.5
        \quad\Longrightarrow\quad
        p\geq 1-0.5^{1/G}\approx 0.083.
    \end{gathered}
    \label{eq:operation-exploration-probability}
\end{equation}
We adopt the nearby rounded threshold $p_{\min}=0.1$, which exceeds this
reference bound.  The regularizer softly constrains normalized operation
probabilities and does not guarantee that every operation appears in each
group.

\section{Training Algorithm}
\label{app:training-algorithm}

\begin{algorithm}[H]
\caption{\textsc{UniSkill} Training Procedure}
\label{alg:uniskill}
\footnotesize
\begin{algorithmic}[1]
\Require Initial policy $\pi_\theta$; skillbank $\mathcal{S}$; task distribution
$\mathcal{D}$; group size $G$; training steps $N$; learning rate $\alpha$
\Ensure Updated policy $\pi_\theta$ and skillbank $\mathcal{S}$
\For{training step $n=1,\ldots,N$}
    \State $(\pi_{\mathrm{old}},\mathcal{S}_{\mathrm{old}})
    \gets(\pi_\theta,\mathcal{S})$;
    $\mathcal{Q}\gets\Call{SampleTasks}{\mathcal{D}}$
    \State $(s_q,\mathcal{T}_q)\gets
    \Call{ActorRollouts}{\pi_{\mathrm{old}},\mathcal{S}_{\mathrm{old}},q,G}$,
    $\forall q\in\mathcal{Q}$
    \State $\mathcal{L}_{\mathrm{actor}}\gets
    \Call{GRPO}{\{\mathcal{T}_q\}_{q\in\mathcal{Q}}}$
    \Comment{Eqs.~\ref{eq:grpo-main}--\ref{eq:grpo-actor-objective}}
    \State $\mathcal{P}\gets
    \Call{GenerateSkillProposals}{\pi_{\mathrm{old}},\{\mathcal{T}_q\}_{q\in\mathcal{Q}}}$
    \Comment{Source and opposite-outcome inputs; Eq.~\ref{eq:skill-proposal}}
    \For{each proposal $p=(\tau,\tau_{\mathrm{opposite}},\mathrm{op},\tilde{s})\in\mathcal{P}$}
        \State $(\tau^+,\tau^-)\gets
        \Call{SampleReferences}{\mathcal{T}_q\setminus\{\tau,\tau_{\mathrm{opposite}}\}}$
        \Comment{Eq.~\ref{eq:input-excluded-references}}
        \State $\mathbf{d}_p\gets\Call{ValidateProposal}{p}$
        \Comment{Format and frozen critic checks}
        \State $(\Delta_p^+,R_{\mathrm{align},p})\gets
        \Call{ContrastiveFeedback}{\pi_{\mathrm{old}},p,\tau^+,\tau^-,\mathbf{d}_p}$
        \Comment{Eqs.~\ref{eq:actor-relative-score}--\ref{eq:r-align}}
        \State $r_{\mathrm{prop},p}\gets
        \Call{ProposalReward}{p,\mathbf{d}_p,R_{\mathrm{align},p}}$
        \Comment{Eqs.~\ref{eq:proposal-structure-operation-reward}--\ref{eq:reinforcepp-proposal-advantage}}
    \EndFor
    \State $\mathcal{L}_{\mathrm{proposer}}\gets
    \Call{REINFORCE++}{\mathcal{P},\{r_{\mathrm{prop},p}\}}+
    \lambda_{\mathrm{sup}}\Call{SkillEditSupport}{\pi_\theta,\mathcal{P}}$
    \Comment{Eqs.~\ref{eq:reinforcepp-proposal-advantage}--\ref{eq:reinforcepp-proposer-objective}}
    \State $\theta\gets\Call{JointUpdate}{\theta,\mathcal{L}_{\mathrm{actor}},
    \mathcal{L}_{\mathrm{proposer}},\alpha}$
    \Comment{Eq.~\ref{eq:joint-objective}}
    \State $\mathcal{S}\gets
    \Call{ApplyAcceptedEdits}{\mathcal{S}_{\mathrm{old}},\mathcal{P},
    \{\mathbf{d}_p,\Delta_p^+,R_{\mathrm{align},p}\}}$
    \Comment{Sec.~\ref{sec:joint-evolution}}
\EndFor
\end{algorithmic}
\end{algorithm}

\section{Additional Experimental Details}
\label{app:additional-experiments}

\subsection{Out-of-Distribution Generalization}
\label{app:ood-generalization}

We evaluate UniSkill with and without skill retrieval on ALFWorld's
\texttt{valid\_unseen} split over three independent training runs.  Within each
run, both settings use the same trained policy, with retrieval drawing on the
skillbank learned in that run.  Neither the policy nor the skillbank is
updated during evaluation.

\begin{table}[H]
    \small
    \caption{UniSkill's ALFWorld out-of-distribution success rates (\%),
    reported as mean $\pm$ sample std over three independent training runs.}
    \label{tab:ood-generalization}
    \begin{center}
    \setlength{\tabcolsep}{3pt}
    \begin{tabular}{lccccccc}
        \toprule
        Setting & Pick & Look & Clean & Heat & Cool & Pick2 & All \\
        \midrule
        w/ Retrieval
        & 100.0$_{\pm 0.0}$ & 100.0$_{\pm 0.0}$ & 94.1$_{\pm 5.9}$
        & 100.0$_{\pm 0.0}$ & 100.0$_{\pm 0.0}$ & 98.2$_{\pm 3.0}$
        & 98.2$_{\pm 1.2}$ \\
        w/o Retrieval
        & 100.0$_{\pm 0.0}$ & 92.9$_{\pm 0.0}$ & 92.2$_{\pm 3.4}$
        & 100.0$_{\pm 0.0}$ & 98.0$_{\pm 3.4}$ & 98.2$_{\pm 3.0}$
        & 96.6$_{\pm 0.9}$ \\
        \bottomrule
    \end{tabular}
    \end{center}
\end{table}

As shown in Table~\ref{tab:ood-generalization}, skill retrieval increases mean
overall success from 96.6\% to 98.2\%, a gain of 1.6 percentage points.  Both
settings perform well on unseen environments, with retrieval providing
additional gains on Look, Clean, and Cool.

\subsection{Sensitivity to the Skill Critic}
\label{app:critic-sensitivity}

\begin{table}[H]
    \small
    \centering
    \caption{WebShop performance with different skill critics.}
    \label{tab:critic-sensitivity}
    \vspace{0.6em}
    \setlength{\tabcolsep}{12pt}
    \begin{tabular}{lcc}
        \toprule
        Skill Critic & Task Score & Success (\%) \\
        \midrule
        DeepSeek-V4-Pro & 90.5\rlap{$_{\pm 1.4}$} & 84.7\rlap{$_{\pm 0.5}$} \\
        DeepSeek-V4-Flash & 88.9 & 83.6 \\
        Qwen3.5-35B-A3B & 90.4 & 85.2 \\
        \bottomrule
    \end{tabular}
\end{table}

\begin{figure}[!htbp]
    \centering
    \includegraphics[width=0.75\linewidth]{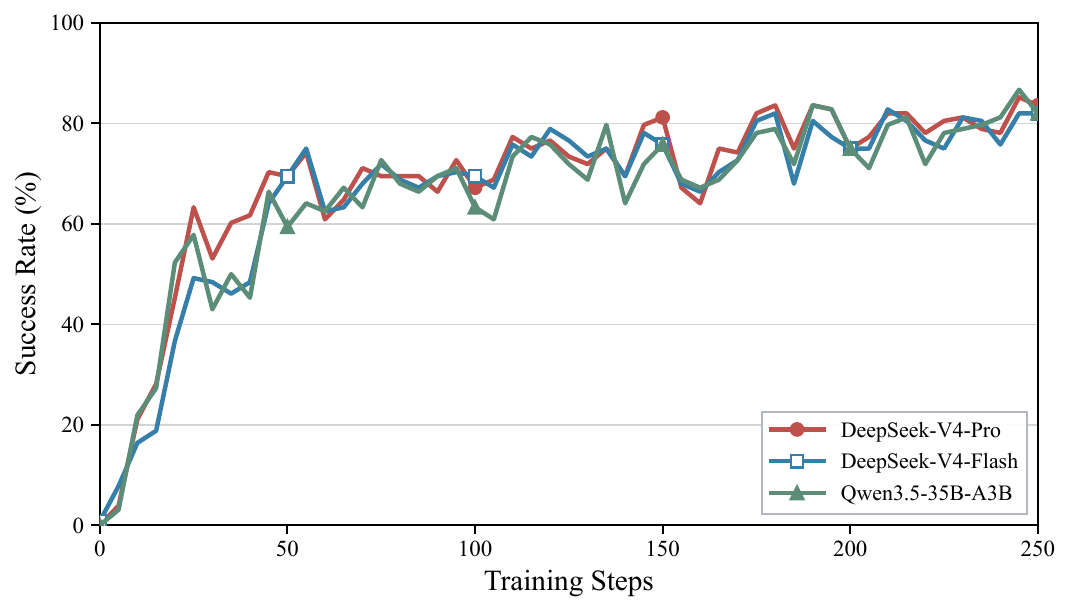}
    \caption{WebShop validation success with different skill critics.}
    \label{fig:critic-model-success-rate}
\end{figure}

We also test DeepSeek-V4-Flash and Qwen3.5-35B-A3B as WebShop skill critics.
Table~\ref{tab:critic-sensitivity} reports the three-run mean and standard
deviation for DeepSeek-V4-Pro and single-run results for the alternatives.
Figure~\ref{fig:critic-model-success-rate} shows similar validation-success
trajectories, alongside comparable final task scores and success rates in the
table.  This is consistent with the critic filtering inappropriate skill-edit
operations and trajectory-unsupported content, while the actor-dependent
$R_{\mathrm{align}}$ provides the alignment signal.  Together, these results
suggest that UniSkill's performance is not tied to the default critic among
those tested.

\subsection{Actor-Alignment Evaluation}
\label{app:actor-alignment-evaluation}

For Q2, we load the model checkpoints saved at training steps 25 and 75
and collect trajectories on randomly sampled queries from the ALFWorld
training split.  Following the training procedure, we generate skill
proposals and retain those for which $R_{\mathrm{align}}$ can be computed.
For the plotted analysis, we randomly sample 100 eligible proposals at each
checkpoint without conditioning on their rollout gains.  We hold the
corresponding model fixed and evaluate
each proposal on its source query with 32 rollouts under the proposed skill
and 32 under the original skill condition.  The difference in success rates
is plotted against $R_{\mathrm{align}}$ in
Figure~\ref{fig:alignment-validation}.

\subsection{Additional Skill-Edit Dynamics}
\label{app:additional-skill-edit-dynamics}

\begin{figure}[!htbp]
    \centering
    \includegraphics[width=\linewidth]{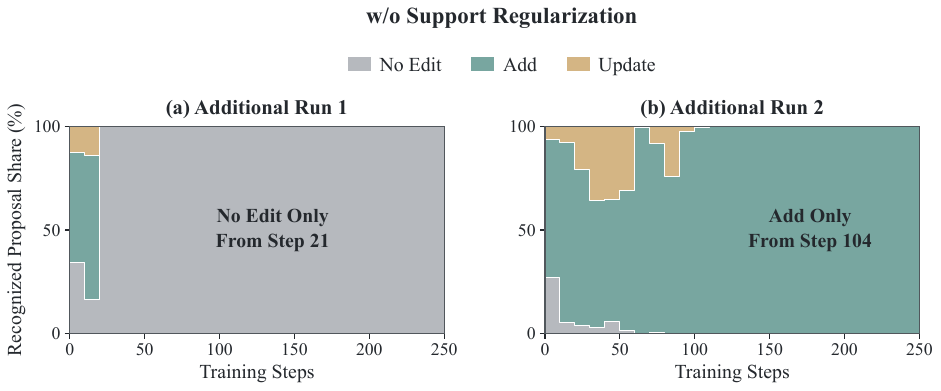}
    \caption{Skill-edit distributions for two additional independent runs
    without support regularization on ALFWorld.  Shares are computed from
    recognized proposals pooled over non-overlapping 10-step windows.}
    \label{fig:additional-skill-edit-dynamics}
\end{figure}

Figure~\ref{fig:additional-skill-edit-dynamics} shows skill-edit distributions
for two additional runs without support regularization.  Both concentrate on
a single operation, but the dominant operation differs: \textsc{No Edit} in
one run and \textsc{Add} in the other.  Together with
Figure~\ref{fig:skill-edit-exploration}, these results show that the observed
collapse is not specific to \textsc{Add} or to a single run.

\clearpage
\section{Case Studies}
\label{app:qualitative-case-studies}

We organize the cases from two complementary viewpoints.  First, we follow
committed skills across training to show how interaction evidence produces
\textsc{Add} and \textsc{Update} operations and how the resulting guidance
is later reused.  Second, we use the Q2 protocol to isolate the immediate
effect of a candidate skill: the task and actor checkpoint are fixed, and
only the directly supplied skill condition changes.  The fixed-actor cases
include cases in which $R_{\mathrm{align}}$ and rollout utility agree,
covering both positive-positive and negative-negative outcomes, as well as
diagnostic cases in which positive alignment does not yield a positive
rollout gain.  We report raw $R_{\mathrm{align}}$ unless clipping is stated.

\tcbset{uniskillcase/.style={
    enhanced, breakable, width=\linewidth,
    colback=black!2, colframe=black!25,
    boxrule=0.45pt, arc=0.7mm,
    left=2.5mm, right=2.5mm, top=2mm, bottom=2mm,
    before skip=12pt, after skip=7pt,
    fonttitle=\small\bfseries,
    before upper={\small\raggedright\setlength{\parindent}{0pt}
    \setlength{\parskip}{3pt}}
}}

\subsection{View I: Skillbank Evolution over Training}
\label{app:case-training-evolution}

These longitudinal cases illustrate how the skillbank is constructed and
reused during training.  Actor-specific utility is examined separately in
the fixed-actor cases in Section~\ref{app:case-fixed-actor}.

\subsubsection{Adding a Missing Search Location}
\label{app:case-kettle-stoveburner}

The actor was asked to place a kettle on a shelf.  The retrieved skill only
suggested searching cabinets, the dining table, and the fridge.  After searches
at these locations failed, the actor eventually found the kettle on a
stove burner, revealing a missing fallback location.

\begin{tcolorbox}[uniskillcase, title={Skill change}]
\begin{tabular}{@{}p{0.16\linewidth}p{0.78\linewidth}@{}}
\textbf{Task} & Put a kettle on a shelf. \\[2pt]
\textbf{Retrieved} & If the item is missing, check cabinets, the dining table,
and the fridge before placing it on the countertop. \\[2pt]
\textbf{Updated} & Generalize the destination to a shelf or other storage
location; after checking countertops and cabinets, also check shelves, the
microwave, and stove burners. \\[2pt]
\textbf{Decision} & Critic accepted; \textsc{Update} committed to the
skillbank. \\
\end{tabular}
\end{tcolorbox}

The critic identified the stove burner as a trajectory-supported location
missing from the retrieved guidance.  The proposal received
$\Delta^+=0.00110$, $\Delta^-=-0.01951$, and therefore
$R_{\mathrm{align}}=0.02061$.  It satisfied both admission
criteria, and the accepted proposal updated the existing skill in the
skillbank.  Later in training, the exact same task instance---with the same
task text and ALFWorld game file, but a new query ID---was sampled again.  It
retrieved the same skill ID, now containing the committed update.

\begin{tcolorbox}[uniskillcase, title={Later reuse on the same task instance}]
\begin{tabular}{@{}p{0.19\linewidth}p{0.75\linewidth}@{}}
\textbf{Initial occurrence} & Eight rollouts used the retrieved skill and
averaged 28.25 interactions; 6/8 succeeded. \\[3pt]
\textbf{Committed update} & Add shelves, the microwave, and stove burners as
fallback search locations. \\[3pt]
\textbf{Later recurrence} & The same task instance retrieved the updated
skill; eight rollouts averaged 16.75 interactions and 8/8 succeeded. \\
\end{tabular}
\end{tcolorbox}

This longitudinal record establishes that the committed update was later
retrieved and reused when the same task instance recurred.  The later
rollouts were shorter and all succeeded.

\Needspace{30\baselineskip}
\subsubsection{Evolving a Shared Lamp-Search Skill}
\label{app:case-desklamp-evolution}

The ALFWorld \texttt{look\_at\_obj\_in\_light} task requires the actor to
obtain a target object and locate and use a desk lamp to examine it.  The
target may change from a bowl to a pen, statue, or pencil, but every task
contains the same lamp-search subproblem.  This case follows a skill that
stores where to search for the desk lamp as successive interactions expose
locations missing from the current guidance.

\begin{table}[H]
    \small
    \centering
    \caption{Evolution of one ALFWorld lamp-search skill.  Each row is a
    critic-accepted edit committed to the skillbank; the last column reports
    raw alignment scores.}
    \label{tab:desklamp-skill-evolution}
    \setlength{\tabcolsep}{4pt}
    \renewcommand{\arraystretch}{1.12}
    \begin{tabular}{@{}cl
        >{\raggedright\arraybackslash}p{0.34\linewidth}
        >{\raggedright\arraybackslash}p{0.29\linewidth}r@{}}
        \toprule
        Edit & Target & Interaction evidence & Key skill change &
        $R_{\mathrm{align}}$ \\
        \midrule
        Add & Bowl & No prior skill; the lamp was found on a dresser. & Add
        the dresser as a lamp-search location. & 0.09850 \\
        Update & Pen & Dresser-only guidance missed a lamp located on a shelf.
        & Add shelves alongside the dresser. & 0.01724 \\
        Update & Bowl & The lamp was found on a side table, which the current
        skill omitted. & Add side tables alongside dressers and shelves.
        & 0.11585 \\
        \bottomrule
    \end{tabular}
\end{table}

The final update obtained $\Delta^+=0.09273$, $\Delta^-=-0.02312$, and
$R_{\mathrm{align}}=0.11585$.  Its training contribution was clipped to
$0.05$, as was the initial Add; the middle Update remained unclipped.

After this update was committed, later task instances of the same
\texttt{look\_at\_obj\_in\_light} type retrieved the same skill ID from the
skillbank.  The target object changed, but the updated lamp-search guidance
was shared across the tasks.

\begin{tcolorbox}[uniskillcase, title={Later retrieval of the updated skill}]
\begin{tabular}{@{}p{0.20\linewidth}p{0.74\linewidth}@{}}
\textbf{Statue task} & A later task instance retrieved the updated skill.  The
actor took the statue from a dresser, moved to a side table, and used the desk
lamp; 8/8 rollouts succeeded. \\[3pt]
\textbf{Pencil task} & Another later task instance retrieved the same updated
skill.  The actor moved to a side table containing both the pencil and desk
lamp, then used the lamp; 8/8 rollouts succeeded. \\
\end{tabular}
\end{tcolorbox}

Thus, ``reuse'' here means retrieval of the same committed skill in subsequent
training tasks, rather than merely observing similar action sequences.  The
skill transfers because these tasks share the lamp-search subproblem even
though their target objects differ.

\Needspace{28\baselineskip}
\subsection{View II: Fixed-Actor Skill Effects}
\label{app:case-fixed-actor}

For each case below, we directly supply the original or candidate skill to
the same actor checkpoint and run 32 rollouts per condition.  This separates
the effect of skill conditioning from changes in actor parameters.

\subsubsection{When Alignment and Rollout Utility Agree}

\medskip
\noindent\textbf{Positive alignment and positive rollout gain.}

\paragraph{Positive \textsc{Add}: finding a desk lamp.}
\label{app:case-alarmclock-fixed-actor}

The task was to examine an alarm clock under a desk lamp.  The actor could
take the alarm clock from a desk, but needed to find the lamp on a dresser
to finish.  With no retrieved skill, it had no explicit lamp-search hint;
the proposed \textsc{Add} named likely lamp locations, including dressers.

\begin{tcolorbox}[uniskillcase, title={Skill change}]
\begin{tabular}{@{}p{0.16\linewidth}p{0.78\linewidth}@{}}
\textbf{Task} & Look at an alarm clock under a desk lamp. \\[2pt]
\textbf{Retrieved} & No skill. \\[2pt]
\textbf{Proposed} & If no desk lamp is available, search likely locations
such as desks, dressers, and drawers; use a found lamp to examine the
object. \\
\end{tabular}
\end{tcolorbox}

The candidate scored $\Delta^+=0.03236$, $\Delta^-=-0.02092$, and raw
$R_{\mathrm{align}}=0.05328$ (clipped to $0.05$ for training).  It meets
the skillbank admission criteria, $\Delta^+>0$ and
$R_{\mathrm{align}}>0$.  To isolate its immediate effect, we supplied
either no skill or the candidate directly to the same fixed actor.

\begin{tcolorbox}[uniskillcase, title={Fixed-actor comparison: 32 rollouts per condition}]
\begin{tabular}{@{}p{0.19\linewidth}p{0.75\linewidth}@{}}
\textbf{No skill} & Reached the dresser and used the desk lamp in 26/32
rollouts; succeeded in 26/32. Mean length: 23.66 interactions. \\[3pt]
\textbf{Candidate} & Reached the dresser and used the desk lamp in 32/32
rollouts; succeeded in 32/32. Mean length: 5.28 interactions. \\
\end{tabular}
\end{tcolorbox}

For example, the no-skill actor took the alarm clock at
interaction~2 but repeatedly examined the desk and clock, never found
the lamp, and timed out at 50 interactions.  With the candidate skill,
it took the clock at~2, went to the dresser at~3, and used the lamp at~4.
Across 32 rollouts, the location hint led the fixed actor to use the lamp
more frequently and substantially earlier.

\Needspace{28\baselineskip}
\paragraph{Positive \textsc{Update}: opening closed cabinets.}
\label{app:case-bowl-fixed-actor}

The task instruction was \texttt{put some bowl on fridge}.  The actor
needed to find a bowl and move it to the fridge; a usable bowl was inside
a closed cabinet.  The retrieved skill mentioned cabinets as possible
search locations, but did not say to open them and was written for a
countertop destination.  The candidate \textsc{Update} generalized the
destination and added an explicit closed-cabinet search step.

\begin{tcolorbox}[uniskillcase, title={Skill change}]
\begin{tabular}{@{}p{0.16\linewidth}p{0.78\linewidth}@{}}
\textbf{Task} & Find a bowl and move it to the fridge. \\[2pt]
\textbf{Retrieved} & For a countertop destination, check cabinets, the
dining table, or the fridge if the item is missing. \\[2pt]
\textbf{Proposed} & Generalize to the specified destination; search
cabinets, countertops, and dining tables; if the item is still missing,
open closed cabinets and continue searching. Once found, place it at the
target. \\
\end{tabular}
\end{tcolorbox}

The proposal received $\Delta^+=0.00805$, $\Delta^-=-0.01391$, and
$R_{\mathrm{align}}=0.02196$ (unclipped), satisfying the skillbank
admission criteria.  We directly supplied the retrieved or
proposed skill to the same actor for 32 rollouts per condition.

\begin{tcolorbox}[uniskillcase, title={Fixed-actor comparison: 32 rollouts per skill}]
\begin{tabular}{@{}p{0.19\linewidth}p{0.75\linewidth}@{}}
\textbf{Retrieved} & Opened a cabinet containing a bowl in 9/32
rollouts; took a bowl in 7/32; succeeded in 7/32. Mean length: 44.53
interactions. \\[3pt]
\textbf{Proposed} & Opened a cabinet containing a bowl in 23/32
rollouts; took a bowl in 21/32; succeeded in 21/32. Mean length: 34.22
interactions. \\
\end{tabular}
\end{tcolorbox}

For example, the retrieved-skill actor reached cabinet~2
at interaction~29 but never opened it and timed out at~50.  With the
proposed skill, it reached the cabinet at~4, opened it at~5, took the
bowl at~6, and completed the task at~9.  The proposed update explicitly
instructed the actor to open closed cabinets.  Across 32 rollouts, the fixed
actor retrieved bowls more often, and task success increased from 7/32 to
21/32.

\Needspace{24\baselineskip}
\medskip
\noindent\textbf{Negative alignment and negative rollout gain.}
\label{app:case-negative-agreement}

The task was to place two newspapers in an armchair.  With no retrieved
skill, the actor could search and place the newspapers directly.

\begin{tcolorbox}[uniskillcase, title={Skill change}]
\begin{tabular}{@{}p{0.16\linewidth}p{0.78\linewidth}@{}}
\textbf{Task} & Place two newspapers in an armchair. \\[2pt]
\textbf{Retrieved} & No skill. \\[2pt]
\textbf{Proposed} & Check likely locations one by one.  Once an object is
found, take it to the target location.  Repeat until all required objects are
found and placed. \\
\end{tabular}
\end{tcolorbox}

The proposal did not explicitly require placing the held object before
restarting the search.  It received $\Delta^+=-0.01375$,
$\Delta^-=0.00277$, and $R_{\mathrm{align}}=-0.01652$.

\begin{tcolorbox}[uniskillcase, title={Fixed-actor comparison: 32 rollouts per condition}]
\begin{tabular}{@{}p{0.19\linewidth}p{0.75\linewidth}@{}}
\textbf{No skill} & Took at least one newspaper in 30/32 rollouts and placed
at least one in 25/32; took two in 25/32 and placed two in 18/32; succeeded
in 17/32. \\[3pt]
\textbf{Candidate} & Took at least one newspaper in 31/32 rollouts but placed
at least one in only 8/32; took two in 7/32 and placed two in 4/32; succeeded
in 4/32. \\
\end{tabular}
\end{tcolorbox}

For example, the no-skill actor placed both newspapers and finished in 17
interactions.  With the candidate skill, it reached the armchair holding the
first newspaper but restarted the search without placing it, moving between
the armchair and side tables until the 50-interaction limit.  This pattern was
common: 23/32 candidate rollouts took a newspaper without placing any,
compared with 5/32 without a skill.  The negative alignment score therefore
correctly identifies guidance whose missing placement-before-search step
reduces closed-loop utility for the current actor.

\Needspace{30\baselineskip}
\subsubsection{When Alignment and Rollout Utility Disagree}
\label{app:case-positive-negative}

Of the 100 proposals evaluated at training step 25 in Q2, 53 have positive
$R_{\mathrm{align}}$, and 13 of these produce a negative rollout gain.  The
skillbank admission criteria exclude 4 because $\Delta^+\leq 0$, leaving 9
that satisfy both $\Delta^+>0$ and $R_{\mathrm{align}}>0$.  We first examine
one excluded proposal to illustrate the $\Delta^+>0$ safeguard.  We then
analyze two admitted proposals that expose complementary residual mismatches
between skill content and actor execution.

\begin{table}[H]
    \small
    \centering
    \caption{Representative cases with positive alignment and negative
    fixed-actor rollout gain.}
    \label{tab:positive-alignment-negative-gain}
    \setlength{\tabcolsep}{3.5pt}
    \renewcommand{\arraystretch}{1.08}
    \begin{tabular*}{\linewidth}{@{\extracolsep{\fill}}
        >{\raggedright\arraybackslash}p{0.31\linewidth}
        rrr>{\centering\arraybackslash}p{0.14\linewidth}
        >{\centering\arraybackslash}p{0.07\linewidth}@{}}
        \toprule
        \multicolumn{1}{c}{Task} &
        \multicolumn{1}{c}{$\Delta^+$} &
        \multicolumn{1}{c}{$\Delta^-$} &
        \multicolumn{1}{c}{$R_{\mathrm{align}}$} &
        \multicolumn{1}{c}{Success} &
        \multicolumn{1}{c}{Admit} \\
        \midrule
        Clean kettle $\rightarrow$ stove burner
            & $-0.00449$ & $-0.01515$ & 0.01067 & 15/32 $\rightarrow$ 8/32 & No \\
        \addlinespace[2pt]
        Two kettles $\rightarrow$ one cabinet
            & 0.00087 & $-0.02682$ & 0.02769 & 30/32 $\rightarrow$ 24/32 & Yes \\
        Two remotes $\rightarrow$ armchair
            & 0.00280 & $-0.01030$ & 0.01311 & 16/32 $\rightarrow$ 6/32 & Yes \\
        \bottomrule
    \end{tabular*}
    \vspace{2pt}

    \parbox{\linewidth}{\footnotesize\emph{Note.} Success reports original
    $\rightarrow$ candidate skill conditions, with 32 rollouts per condition.
    Admit indicates whether the skillbank admission criteria are satisfied.}
\end{table}

Table~\ref{tab:positive-alignment-negative-gain} gives the numerical overview.
We unpack its three rows below in the same order.  For each case, the gray box
states the relevant skill change, why the alignment score is positive, and
what changes in the 32 new rollouts.  The interpretation below the box then
identifies the source of the disagreement.

\medskip
\noindent\textbf{Case 1: Search expansion and the admission safeguard.}
\par\vspace{5pt}
\begin{tcolorbox}[uniskillcase,
    title={Task: Clean a kettle; place it on a stove burner},
    before skip=0pt, left=8pt, right=8pt, top=5pt, bottom=5pt]
\textbf{Skill change.}
The retrieved skill prioritizes countertops, the sink basin, and cabinets
when searching for the object.  The candidate broadens this list to include
the dining table and storeroom, and calls for a more systematic fallback
search.  The intended correction is to reduce failures caused by an
overly narrow search.

\smallskip
\textbf{Why $R_{\mathrm{align}}>0$.}
The candidate lowers both reference log-likelihoods, but lowers the failed
reference more: $\Delta^+=-0.00449$ and $\Delta^-=-0.01515$.  Their difference
is therefore positive, $R_{\mathrm{align}}=0.01067$.  Crucially,
$\Delta^+<0$, so the proposal does not pass the skillbank admission
criteria.

\smallskip
\textbf{Behavior in new rollouts.}
The kettle is in a cabinet, which already appears in the retrieved skill's
priority list.  Under the broader candidate, the actor more often spends its
budget revisiting the added dining-table location, countertops, and multiple
cabinets without retrieving the kettle.  Whenever the candidate-conditioned
actor does retrieve the kettle, it completes the remaining cleaning and
placement steps in all 8 cases.

\smallskip
\textbf{Fixed-actor evidence.}
Kettle retrieved: 16/32 $\rightarrow$ 8/32; kettle cleaned and placed:
15/32 $\rightarrow$ 8/32; success: 15/32 $\rightarrow$ 8/32; mean length:
38.69 $\rightarrow$ 43.78.
\end{tcolorbox}

\noindent\textbf{Interpretation.}
The proposal makes a reasonable attempt to improve search coverage, but the
additional locations diffuse this actor's search on the current instance.
The negative $\Delta^+$ detects that the candidate does not better support the
successful reference, even though its relative alignment score is positive.
This is precisely the class of disagreement removed by requiring both
$\Delta^+>0$ and $R_{\mathrm{align}}>0$ before a skillbank update.

The remaining two proposals expose different sources of disagreement.  In
Case~2, the candidate gives a reasonable procedure that the fixed actor does
not follow consistently.  In Case~3, the candidate itself leaves a critical
ordering constraint implicit.

\Needspace{27\baselineskip}
\medskip
\noindent\textbf{Case 2: Multi-object guidance and execution efficiency.}
\par\vspace{5pt}
\begin{tcolorbox}[uniskillcase,
    title={Task: Put two kettles in one cabinet},
    before skip=0pt, left=8pt, right=8pt, top=5pt, bottom=5pt]
\textbf{Candidate guidance.}
No skill is initially retrieved, so the actor relies on its learned policy.
The candidate adds a sensible multi-object procedure in which the actor moves
each object to the target once it is found and searches elsewhere only when a
required object is missing.

\smallskip
\textbf{Why $R_{\mathrm{align}}>0$.}
The multi-object procedure slightly raises the successful-reference
log-likelihood ($\Delta^+=0.00087$) and more strongly lowers the
failed-reference log-likelihood ($\Delta^-=-0.02682$), yielding
$R_{\mathrm{align}}=0.02769$.  The proposal therefore passes the skillbank
admission criteria.

\smallskip
\textbf{Behavior in new rollouts.}
In this task instance, both kettles are visible on the same countertop.
Without a skill, the actor usually follows a compact plan by placing one
kettle, returning to the known location of the second, and reusing the same
cabinet.  Under the candidate condition, some rollouts explore additional
locations even though both kettles have already been observed, while others
place the two kettles in different cabinets.

\smallskip
\textbf{Fixed-actor evidence.}
Three candidate rollouts end without completing two placements, and five
place the kettles in different cabinets.  All eight candidate failures reach
the 50-interaction limit.  Success decreases from 30/32 to 24/32, while mean
length increases from 21.78 to 34.03 interactions.
\end{tcolorbox}

\noindent\textbf{Interpretation.}
The candidate already states that a found object should be moved to the target
and that the search should expand only when a required object is missing.  The
negative gain therefore reflects incomplete alignment between the fixed actor
and the newly proposed skill.  The actor does not consistently follow the
complete guidance, which disrupts the shorter routine used without a skill.

\Needspace{22\baselineskip}
\medskip
\noindent\textbf{Case 3: Multi-object guidance and action ordering.}
\par\vspace{5pt}
\begin{tcolorbox}[uniskillcase,
    title={Task: Put two remote controls in an armchair},
    before skip=0pt, left=8pt, right=8pt, top=5pt, bottom=5pt]
\textbf{Candidate guidance.}
No skill is initially retrieved, so the actor relies on its learned policy.
The candidate states that once one object is found, the actor should use a
loop to find the remaining objects across possible locations.  It also says
that a found object should be moved to the target, but it does not explicitly
require the object already being carried to be placed before the next search
begins.

\smallskip
\textbf{Why $R_{\mathrm{align}}>0$.}
Continuing the search after finding one remote directly matches the task's
two-object requirement.  This guidance slightly raises the
successful-reference log-likelihood ($\Delta^+=0.00280$) and more strongly
lowers the failed-reference log-likelihood ($\Delta^-=-0.01030$), widening
the gap by $R_{\mathrm{align}}=0.01311$.

\smallskip
\textbf{Behavior in new rollouts.}
Without a skill, the actor more often places the first remote before resuming
the search.  In several candidate-conditioned rollouts, the actor reaches the
armchair while holding the first remote control but begins searching for the
second remote before placing the first one.

\smallskip
\textbf{Fixed-actor evidence.}
At least one remote placed: 27/32 $\rightarrow$ 16/32; second remote retrieved:
23/32 $\rightarrow$ 15/32; success: 16/32 $\rightarrow$ 6/32; mean length:
39.03 $\rightarrow$ 45.78 interactions.
\end{tcolorbox}

\noindent\textbf{Interpretation.}
The candidate captures the task's high-level objective, but it leaves the
placement-before-search constraint implicit.  The actor follows the newly
introduced search loop while still holding the first remote.  Because
ALFWorld permits carrying only one object at a time, it cannot retrieve the
second remote until the first has been placed, which lowers rollout utility.

Together, the cases separate three sources of disagreement.  In Case~1, the
additional $\Delta^+>0$ requirement excludes a proposal whose positive
$R_{\mathrm{align}}$ comes from a larger decrease on the failed reference.
Case~2 reveals incomplete alignment between the actor and an otherwise
reasonable new skill, whereas Case~3 reveals an ordering constraint omitted by
the proposal itself.  These complementary failures motivate both sides of
co-evolution.  Continued training can improve the actor's ability to use new
guidance, while subsequent interaction can expose missing constraints and
support further skill refinement.  Such disagreements are relatively limited
and are less frequent at the later checkpoint.  At training step 25, 13 of 53
positive-score proposals have negative rollout gains, and the $\Delta^+>0$
criterion excludes four of them, leaving nine among the 100 evaluated
proposals.  At training step 75, the positive-score disagreement rate is lower
at 10/64 (15.6\%), compared with 13/53 (24.5\%) at training step 25.

\end{document}